%% file: arxiv-version-May.tex
\documentclass{article}

 \usepackage[preprint]{neurips_2026}

\usepackage[utf8]{inputenc} 
\usepackage[T1]{fontenc}    
\usepackage{hyperref}       
\usepackage{url}            
\usepackage{booktabs}       
\usepackage{amsfonts}       
\usepackage{nicefrac}       
\usepackage{microtype}      
\usepackage{xcolor}         

\usepackage{graphicx}
\usepackage{amsmath}
\usepackage{subcaption}
\usepackage{multirow}
\usepackage{booktabs}
\usepackage{paralist}
\usepackage{placeins}
\usepackage[export]{adjustbox}

\def\our{CEDAR}

\title{Conceptualizing Embeddings: Sparse Disentanglement for Vision–Language Models}

\author{
    Piotr Kubaty$^*$ $^{1,2}$ Patryk Marsza{\l}ek$^*$ $^{1,2}$ {\L}ukasz Struski $^{1}$
    \And Adam Wróbel $^{1,2}$ Jacek Tabor $^{1,3}$ Marek \'Smieja $^{1}$ \\ \\
    \textsuperscript{1} Faculty of Mathematics and Computer Science, Jagiellonian University, Kraków, Poland \\
    \textsuperscript{2} Doctoral School of Exact and Natural Sciences, Jagiellonian University, Kraków, Poland \\
    \textsuperscript{3} Centre for Credible AI, Warsaw University of Technology, Warsaw, Poland \\
}

\begin{document}

\maketitle

\input{sections/00_abstract}

\section{Introduction}


Vision–language models such as CLIP~\cite{radford2021learning}, BLIP~\cite{li2022blip}, and CoCa~\cite{yu2022cocacontrastivecaptionersimagetext} are a cornerstone of multimodal learning, achieving strong performance by learning high-dimensional shared embeddings. Despite this success, these representations remain largely opaque: it is unclear how semantic factors are encoded, how they interact, and which components are most relevant for a given input.


Existing interpretability methods capture only a partial view of this structure. Many focus on local explanations, such as highlighting image regions or attributing predictions to features~\cite{sundararajan2017axiomatic, bach2015pixel, abnar2020quantifying, chefer2021transformer, wrobel2026dave}, without revealing the global organization of the embedding space. Others aim to align embeddings with human-interpretable concepts~\cite{koh2020concept, ghorbani2019towards, kim2018interpretability}, but typically rely on predefined directions, supervision,  or assumptions about the original embedding geometry, limiting their flexibility and scalability.


A prominent line of work employs sparse autoencoders (SAEs)~\cite{bricken2023monosemantic, cunningham2024scaling, gao2025topk, rajamanoharan2024gated, zaigrajew2025interpreting} to learn overcomplete representations where neurons correspond to interpretable features. These methods demonstrate that pretrained embeddings contain extractable semantic structure. However, they typically expand the representation into a higher-dimensional space, which alters the original geometry and produces lossy, not directly invertible representations. This makes it difficult to faithfully relate the explanations back to the underlying model.

In this work, we take a different approach. Instead of expanding the representation, we ask: \emph{Can we reveal the semantic structure of pretrained embeddings by reorganizing them within their original space?}

To address this, we propose \our{} (Conceptual Embedding Disentanglement via Adaptive Rotation), a post-hoc method that learns an invertible reparameterization of the embedding space. The key idea is to identify a global coordinate system in which semantic information becomes concentrated in a small number of dimensions. This is achieved by applying a learned transformation and enforcing a top-k sparsity bottleneck in the transformed space, while reconstructing the original embedding.

This formulation leads to a representation with three desirable properties:
\begin{compactitem}
\item Sparsity: each input is described by a small number of active coordinates,
\item Disentanglement: coordinates correspond to approximately independent semantic factors,
\item Axis alignment: each dimension captures a coherent and reusable concept.
\end{compactitem}
Crucially, because the transformation is invertible, \our{} preserves the structure of the original embedding space and enables faithful reconstruction, in contrast to expansion-based methods such as SAEs.


Our approach is model-agnostic and supports different modes of interpretation depending on the vision–language model. In CLIP-like architectures, individual coordinates can be aligned with textual concepts, yielding compact explanations analogous to SAE methods—but without representation expansion. For generative models such as BLIP, these same sparse coordinates can be decoded into natural language descriptions, providing sentence-level explanations. This flexibility enables \our{} to unify concept-based and free-form interpretability within a unified framework.


We evaluate \our{} against several sparse autoencoder variants on standard benchmarks. Our method achieves a competitive reconstruction–sparsity trade-off, particularly in low-dimensional regimes, while yielding more interpretable explanations as demonstrated in a user study. These findings suggest that the semantic structure of vision–language embeddings can be effectively uncovered through a suitable change of basis, without increasing dimensionality or re-training model weights.

\input{sections/02_related_works}


\section{Disentangled sparse reparameterization of embedding space}

In this section, we introduce \our{} (Conceptual Embedding Disentanglement via Adaptive Rotation), a post-hoc method for extracting interpretable structures from frozen vision-language embeddings. Unlike standard fine-tuning, \our{} learns an invertible reparameterization of the embedding space. This approach, inspired by EPIC~\cite{borycki2026epic} and PluGeN~\cite{wolczyk2022plugen}, reorganizes existing features into latent semantic axes without altering the underlying encoder. This way, we effectively \emph{reorganize} existing representations rather than learning new ones.

First, we present the motivation for the problem under consideration. Next, we explain our method and describe its training procedure. Finally, we cover interpretability mechanism for multi-modal architectures.  


\paragraph{Motivation and problem statement.}
\label{sec:method-motivation}

Our core hypothesis is that pretrained embeddings already contain disentangled semantic factors, but these factors are distributed across many coordinates and are therefore difficult to interpret. We seek a coordinate transformation $\mathcal{U}$ that reorganizes the embedding space such that semantic information becomes concentrated in a small number of dimensions. More specifically, we aim to obtain a representation with the following properties:
\begin{compactitem}
    \item \emph{Energy concentration} -- most of the signal mass is captured by the top-$k$ coordinates, allowing each sample to be represented by only a small set of active factors,
    
    \item \emph{Axis alignment} -- individual coordinates correspond to coherent and reusable semantic factors that remain consistent across the data distribution,
    
    \item \emph{Competition} -- dimensions compete for activation, discouraging redundant features and promoting specialization.
\end{compactitem}

Let $\mathbf{z} = \mathcal{E}(x) \in \mathbb{R}^D$ be a fixed embedding obtained from a pretrained image encoder $\mathcal{E}$. We assume the existence of a representation basis in which semantic factors are sparse and approximately axis-aligned. Therefore, we look for $k$ scalars $\alpha_i$ such that the transformed embedding $\tilde{\mathbf{z}} = \mathcal{U}(\mathbf{z})$ is expressed as a sparse decomposition:

\[
\tilde{\mathbf{z}} \approx \sum_{i \in \mathcal{S}} \alpha_i \mathbf{e}_i, \quad |\mathcal{S}| = k, \quad \mathcal{S} \subset \{1, \ldots, D\},
\]

where $\mathcal{S}$ denotes the indices of the top-$k$ coordinates and ${\mathbf{e}_i}$ is the canonical basis. Intuitively, $\mathcal{S}$ represents a small set of disentangled concepts sufficient to reconstruct the original embedding.

Sparse representations are easier to interpret as individual coordinates can be associated with distinct concepts. At the same time, the transformation should remain invertible, enabling faithful reconstruction while preserving the geometric structure of the pretrained space, which is essential for CLIP-like models where text and image embeddings must remain aligned. This allows us to impose sparsity and disentanglement without altering the underlying representation space. This leads us to the following optimization problem:

\emph{Problem formulation.}
Let $\mathbf{z} = \mathcal{E}(x) \in \mathbb{R}^D$. We seek a representation $\hat{\mathbf{z}}$ with bounded sparsity, such that reconstruction error is minimized:
\[
\hat{\mathbf{z}} = \arg\min_{\mathbf{z}'} \|\mathbf{z} - \mathbf{z}'\|_1 \quad \text{s.t.} \quad \|\mathbf{z}'\|_0 \le k.
\]
where $k \ll D$ and $\|.\|_0$ denotes $l_0$ norm

\paragraph{\our{}.}
\begin{figure}[t!]
\centering
\includegraphics[width=\textwidth]{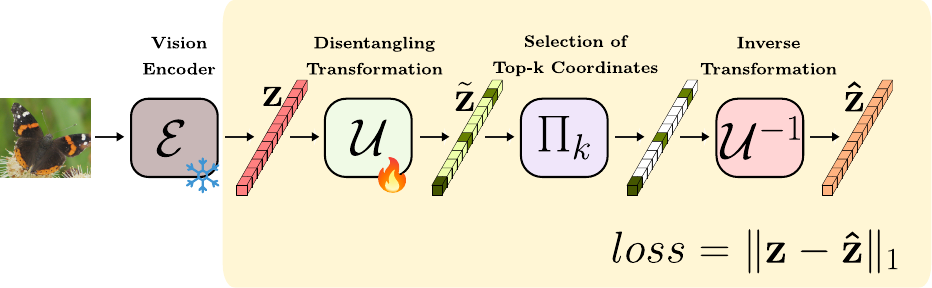}
\caption{Overview of \our{} architecture and training procedure. Given embeddings from a frozen backbone encoder, we apply an invertible, dimension-preserving transformation $\mathcal{U}$ to obtain a proxy representation in which most of the embedding norm is concentrated in a small subset of coordinates. A sparse, compressed representation is then obtained via a top-$k$ selection operator. The inverse transformation $\mathcal{U}^{-1}$ maps the representation back to the original space. The model is trained in a reconstruction manner.}
\label{fig:architecture}
\end{figure}


Rather than solving this sparse approximation problem directly in the original embedding space, where semantic factors are highly entangled, we transform the representation into a space where information becomes more concentrated and axis-aligned. To this end, we first introduce an invertible transformation $\mathcal{U}: \mathbb{R}^D \xrightarrow[\text{onto}]{\text{1-1}} \mathbb{R}^D$ given by:
\[
\tilde{\mathbf{z}} = \mathcal{U}(\mathbf{z}) = U(\mathbf{z} - \mathbf{b}),
\]
where $\mathbf{b}$ denotes the mean embedding computed over the training set, and $U$ is described at the end of this secition. 

We then perform sparsification in the transformed space using a top-$k$ operator $\Pi_k$, which retains the $k$ coordinates with the largest absolute magnitude and sets all remaining coordinates to zero. The resulting sparse representation is mapped back to the original space using the inverse transformation. Formally, we optimize the following reconstruction objective:

\begin{equation}
\min_{U} \; \mathbb{E}_{\mathbf{z}} \left[
\left\|
\mathbf{z} -
\left(
U^{-1}\Pi_k(U(\mathbf{z}-\mathbf{b})) + \mathbf{b}
\right)
\right\|_1
\right].
\label{eq:reconstruction}
\end{equation}

Because $\mathcal{U}$ is invertible, the original embedding can be faithfully reconstructed while preserving the geometric structure of the pretrained representation space.


We parameterize $U$ as an orthogonal linear operator:
\[
U = \exp(A - A^\top),
\]
which guarantees invertibility $U^\top U = I$ and preserves the geometry of the embedding space $\|U\mathbf{z}\|_2 = \|\mathbf{z}\|_2$. The orthogonality constraint prevents the model from trivially concentrating information through simple coordinate scaling or dimensional collapse. Instead, the only way to minimize reconstruction error under top-$k$ sparsification is to redistribute signal energy across coordinates such that most of the information becomes concentrated within a small subset of dimensions.

\paragraph{Training procedure.}

Let
$
P = \{\mathbf{p}_n\}_{n=1}^{N} \subset \mathbb{R}^D$ denote the set of visual embeddings extracted from a frozen pretrained backbone. Depending on the architecture, $P$ may correspond either to a single global representation ($N=1$), as in CLIP-like models, or to a sequence of visual tokens, as in vision encoder--text decoder architectures such as BLIP or CoCa.

Each embedding vector is processed independently, $\hat{\mathbf{p}}_n = U^{-1}\Pi_k(U(\mathbf{p}_n-\mathbf{b}))+\mathbf{b}$ using the proposed orthogonal transformation $U$ and top-$k$ operator. We optimize the average reconstruction error according to Eq. $\ref{eq:reconstruction}$ over all tokens.


A key difficulty arises from the non-smooth transition between dense and sparse representations induced by the $\Pi_k$ operator. To address this, we employ a curriculum over the sparsity level $k$, starting from relatively dense reconstructions and gradually transitioning to the target sparse regime. After reaching the target sparsity, $k$ is randomly sampled during training, which improves robustness across sparsity levels and mitigates overfitting to a fixed $k$ value. Full formulation is provided in Appendix~\ref{app:training_curriculum}.



\paragraph{Sparse semantic explanations in VLMs.}
Once $U$ has been trained, each coordinate in the transformed space corresponds to a direction in the original embedding space, given by $\mathbf{c}_d = U^{-1}_{[d,:]}.$
These directions define a global set of semantic axes shared across all samples.

\emph{Concept-based explanations.} To associate individual axes with human-interpretable concepts, we leverage the CLIP~\cite{radford2021learning} text encoder. Given a predefined vocabulary of text embeddings $\{\mathbf{t}_j\}$, each axis is matched to its nearest semantic concept using cosine similarity:
\[
j^*(d) = \arg\max_j
\frac{
\langle \mathbf{c}_d, \mathbf{t}_j \rangle
}{
\|\mathbf{c}_d\| \|\mathbf{t}_j\|
}.
\]

For a given input, the sparse support set $\mathcal{S}$ identifies the subset of active concepts, while the corresponding coefficients determine their relative contribution:
$
\tilde{\mathbf{z}}
\approx
\sum_{i \in \mathcal{S}} \alpha_i \mathbf{e}_i.
$
This yields a sparse and globally consistent explanation mechanism, where interpretations are expressed directly in terms of reusable semantic coordinates rather than sample-specific attribution maps.

\emph{Free-form explanations.} In generative vision--language models, the same sparse representation can additionally be interpreted through the text decoder. After sparsification and reconstruction, the resulting embeddings are passed to the decoder without modifying the original generation pipeline. Importantly, all visual tokens are reconstructed using the same sparse coordinate system and the same global set of semantic axes.

The generated description therefore depends only on a small subset of active coordinates determined by the sparsity level $k$. By decreasing the sampling temperature and generating multiple captions from the same sparse representation, one can further identify recurring words and semantic patterns. This provides a complementary form of interpretation, where semantic information is expressed not only through aligned coordinates, but also through natural language descriptions induced by the sparse embedding representation.


\section{Interpreting image embeddings with text concepts}

In this section, we first showcase qualitative visual examples of our method, followed by the results of three tasks from the user study we performed. The experimental setup is the same as in Section~\ref{sec:bench}.

\paragraph{Illustrative examples.}
\label{sec:ill_examples}

\begin{figure}[t!]
\centering

\begin{subfigure}{0.49\textwidth}
\caption*{\our{}}
\includegraphics[width=\linewidth, valign=t]{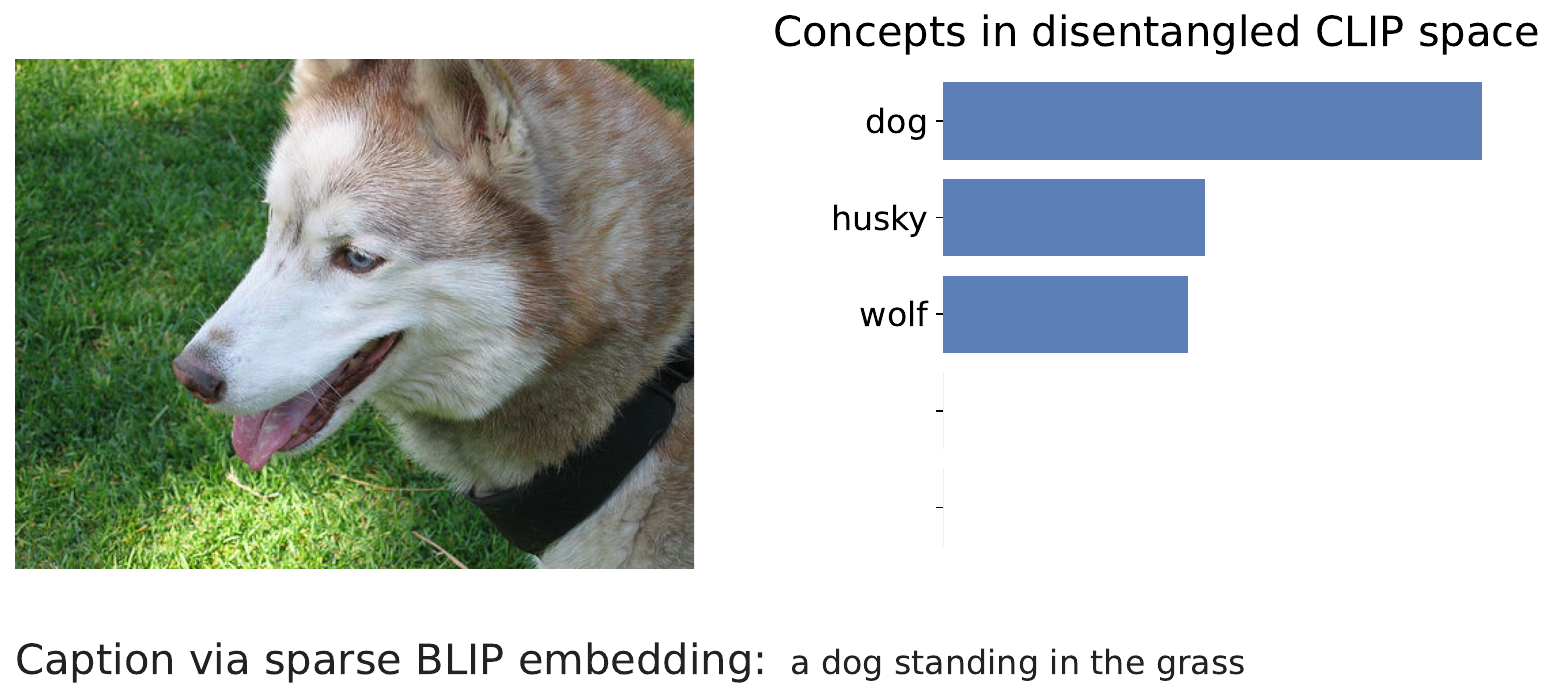}
\end{subfigure}
\hfill
\begin{subfigure}{0.49\textwidth}
\caption*{MSAE}
\includegraphics[width=\linewidth, valign=t]{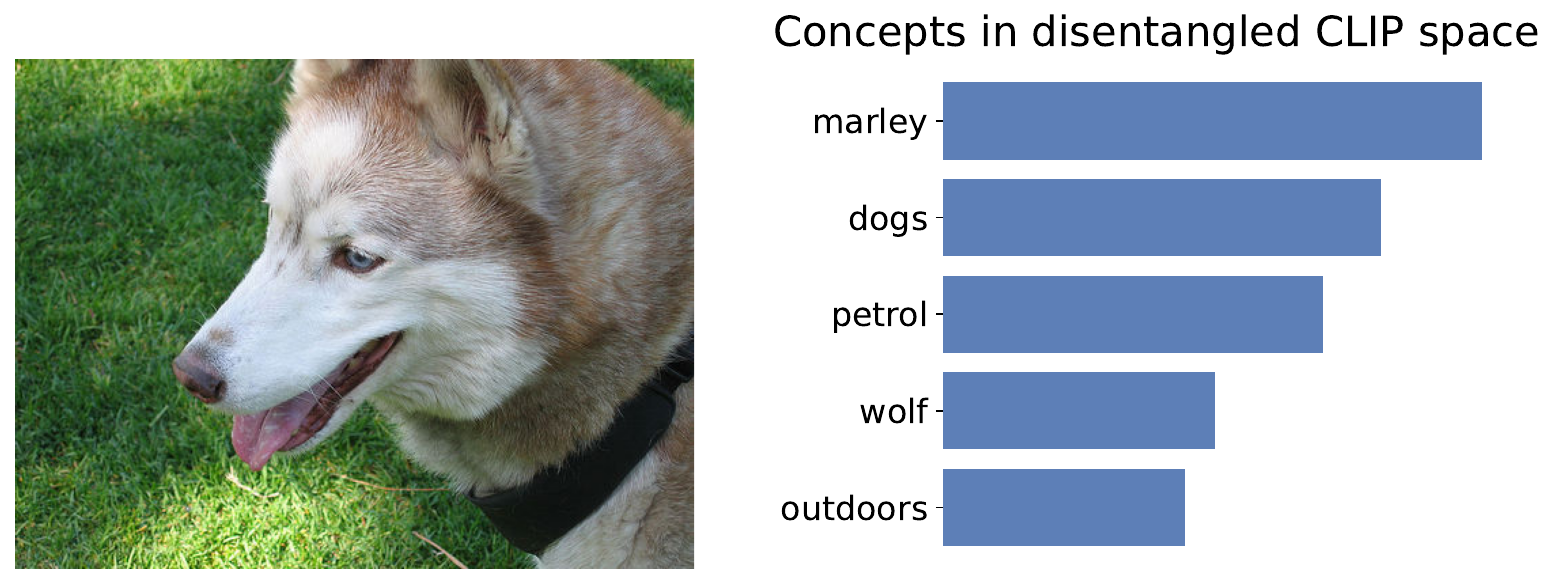}
\end{subfigure}

\begin{subfigure}[t]{0.49\textwidth}
\includegraphics[width=\linewidth, valign=t]{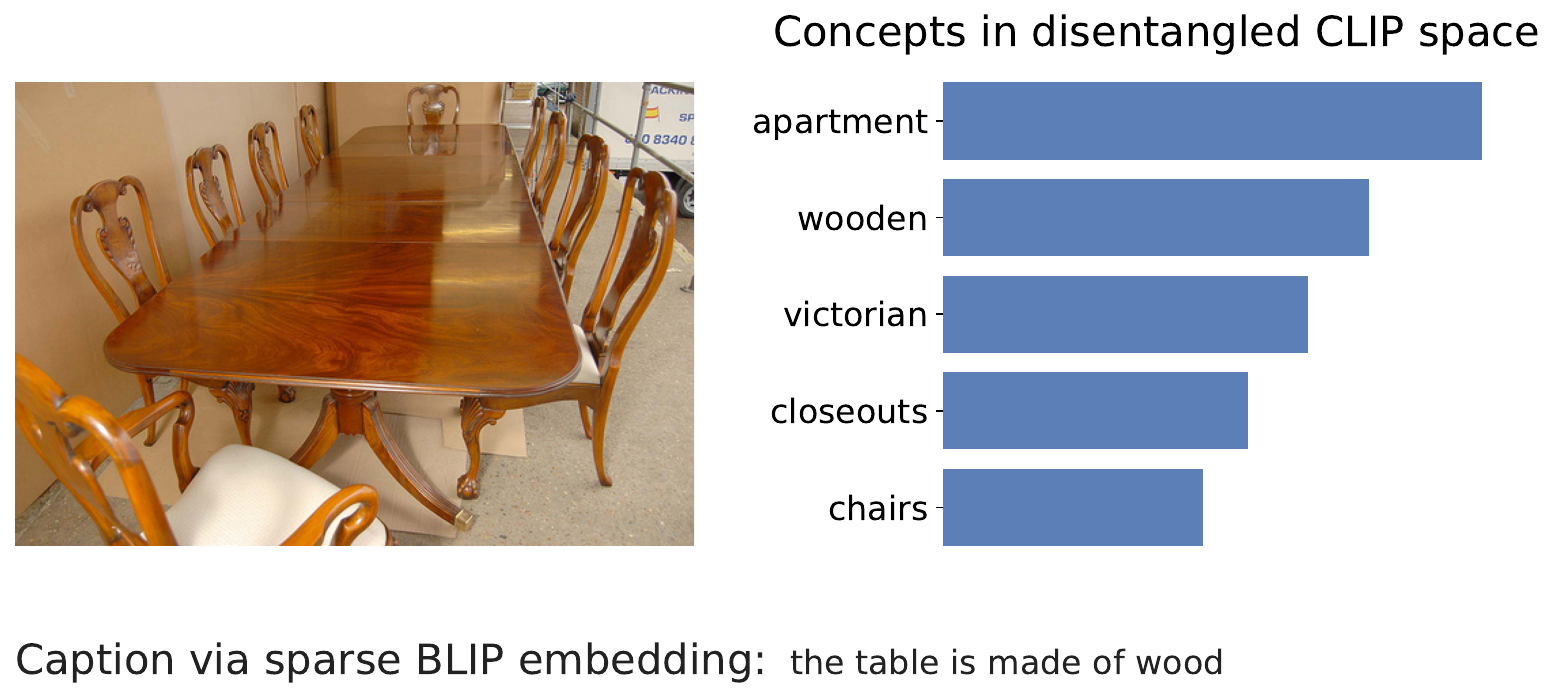}
\end{subfigure}
\hfill
\begin{subfigure}[t]{0.49\textwidth}
\includegraphics[width=\linewidth, valign=t]{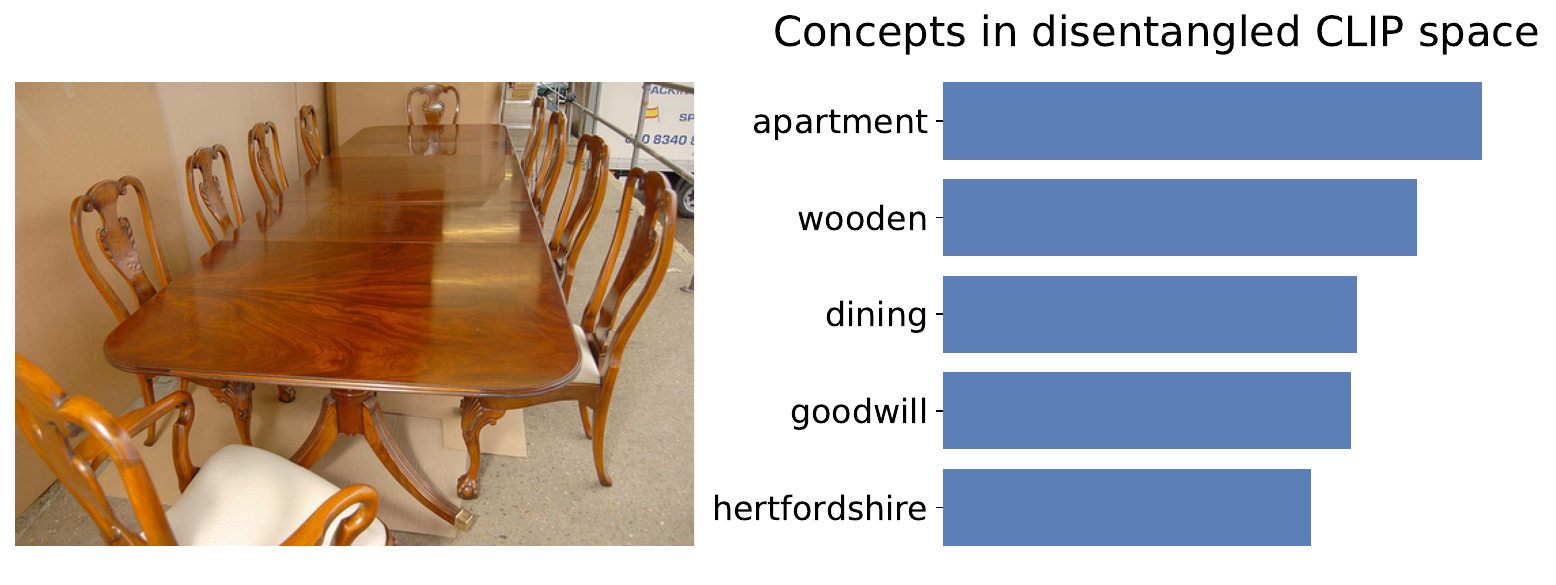}
\end{subfigure}

\caption{Comparison of \our{} and MSAE interpretations. \our{} associates concept keywords with sparse BLIP captions, whereas MSAE is restricted to keyword-based extraction. Bar lengths indicate activation score magnitudes. Notably, in the first example, CEDAR requires fewer active concepts for reconstruction than MSAE.}
\label{fig:text_and_concepts}
\end{figure}

\begin{figure}
    \centering
    \includegraphics[width=\linewidth]{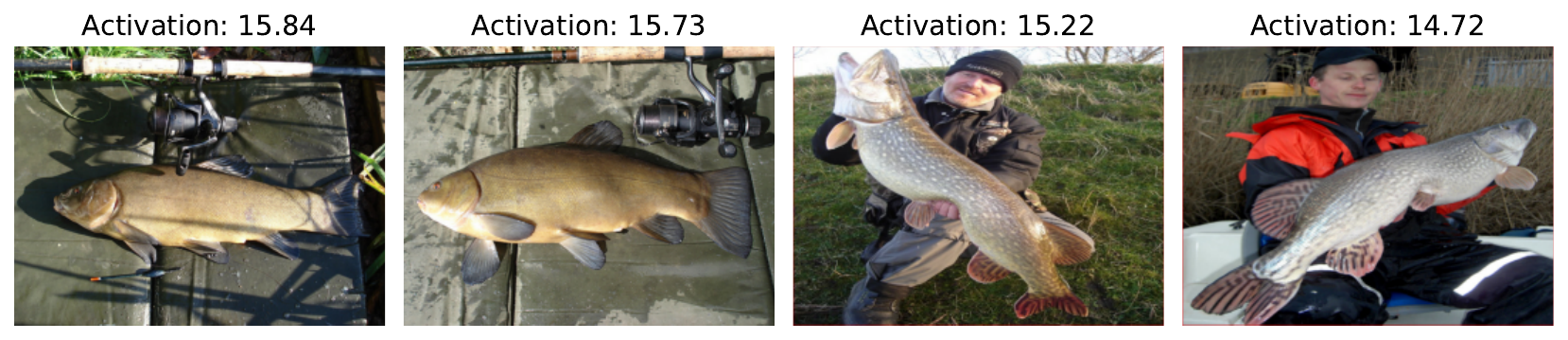}
    \caption{Images with highest activation for neuron 347 in the disentangled space, associated with the concept ``lure''.}
    \label{fig:lure_images}
\end{figure}


We present a visual analysis of how \our{} interprets complex scenes. This includes captions derived from a sparse BLIP~\cite{li2022blip} representation restricted to the top-10 most activating channels as well as disentangled CLIP~\cite{radford2021learning} concepts. We compare these results with those obtained using the Matryoshka Sparse Autoencoder (MSAE)~\cite{zaigrajew2025interpreting}, a recent state-of-the-art SAE model. Figure~\ref{fig:text_and_concepts} illustrates two representative examples where our model bridges the gap between high-level semantics and granular visual attributes. Additional examples showcasing the model's interpretability across various domains are available in Appendix~\ref{app:additional_results}.

In the first example, while MSAE retrieves lower-level or noisier concepts such as “marley” and “petrol,” \our{} more accurately identifies the presence of a “dog” alongside finer-grained taxonomic distinctions, including “husky” and “wolf.” Furthermore, \our{} generates a sparse BLIP caption (“a dog standing on the grass”), demonstrating that semantic fidelity is preserved even under sparse representations, whereas the default MSAE formulation is not capable of producing natural language descriptions. In the second case, \our{} further demonstrates a holistic interpretation by associating stylistic attributes such as “wooden” and “victorian” with a coherent descriptive caption. Together, these results suggest that \our{} effectively disentangles distinct semantic factors, providing a more interpretable structure of the latent space.

Following the interpretation of individual scenes, we analyze the semantic consistency of the disentangled space by examining top activations for specific neurons. Figure~\ref{fig:lure_images} displays the images that most strongly activate Neuron 347, which the model associates with the concept "lure." Across varying contexts, ranging from close shots of caught fish to images of anglers with equipment, the neuron consistently fires for visual features related to fishing. This uniformity indicates that the disentangled dimensions correspond to stable, human interpretable visual concepts rather than arbitrary features.

\paragraph{User study.}
We conducted three user studies to evaluate the quality of concept-based explanations and textual descriptions, including comparisons with MSAE and the dense (non-compressed) model. Additional details of the study protocols and example tasks are provided in Appendix~\ref{app:user-study-details} and Appendix~\ref{app:user-study-examples}. 

In Studies 1 and 2, we compare \our{} with MSAE using embeddings obtained from the CoCa backbone~\cite{yu2022cocacontrastivecaptionersimagetext}. In Study 3, we evaluate an encoder--decoder architecture consisting of a SWIN vision encoder~\cite{liu2021swintransformerhierarchicalvision} and a GPT-2 text decoder~\cite{Radford2019LanguageMA}.

\begin{figure}[htbp]
    \centering
    \includegraphics[width=\textwidth]{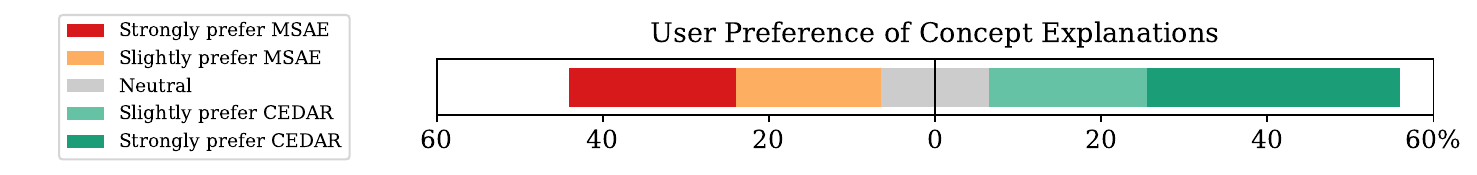}
    \caption{Participants consistently favor \our{} across preference levels, with a significant share expressing a strong preference for our method.}
    \label{fig:user_preference}
\end{figure}

\emph{Study 1: Method preference.} We evaluate pairwise preference between the sets of concept-based explanations using a 5-point scale. As shown in Figure~\ref{fig:user_preference}, \our{} is preferred over MSAE at both intensity levels, indicating a consistent shift towards \our{} rather than isolated strong preferences. This suggests that \our{} produces more representative and useful concept sets for describing image content.

\begin{figure}[htbp]
    \centering
    \includegraphics[width=\textwidth]{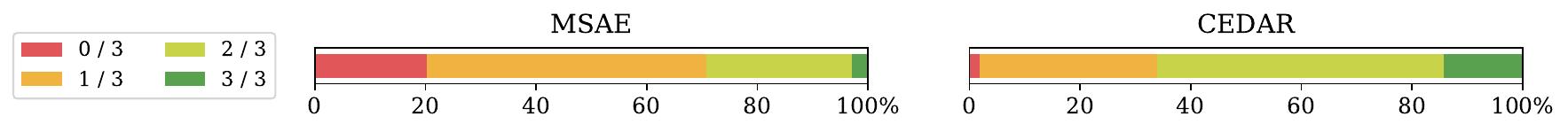}
    \caption{Distribution of the number of  concepts generated by each method being selected. Concepts from \our{} are more frequently identified as relevant compared to MSAE.}
    \label{fig:selection_distribution}
\end{figure}

\emph{Study 2: Concept selection.} We assess how individual concepts align with human judgments by asking participants to select all relevant concepts from a mixed set comprising 3 elements per method. Analysis of the selection distribution, shown in Figure~\ref{fig:selection_distribution}, reveals a clear advantage of \our{} over MSAE. Specifically, \our{} exhibits a significantly higher density in the top-tier categories, where a majority (2 or 3 out of 3) of its suggested concepts were selected. Conversely, MSAE's distribution is skewed toward the lower-end, with most cases 0 or 1 concepts being selected. This shift indicates that \our{} consistently provides a more relevant set of descriptors, whereas MSAE's concepts are frequently rejected by human evaluators.

\emph{Study 3: Description quality.}
We evaluate the quality of full textual descriptions obtained with text decoder based on the representations created by \our{}, a sparsified representation w/o disentanglement,s and full dense representation (no sparsification) using a 3-point scale: does not describe, partially describes, and describes the image.

\begin{figure}[htbp]
    \centering
    \includegraphics[width=\textwidth]{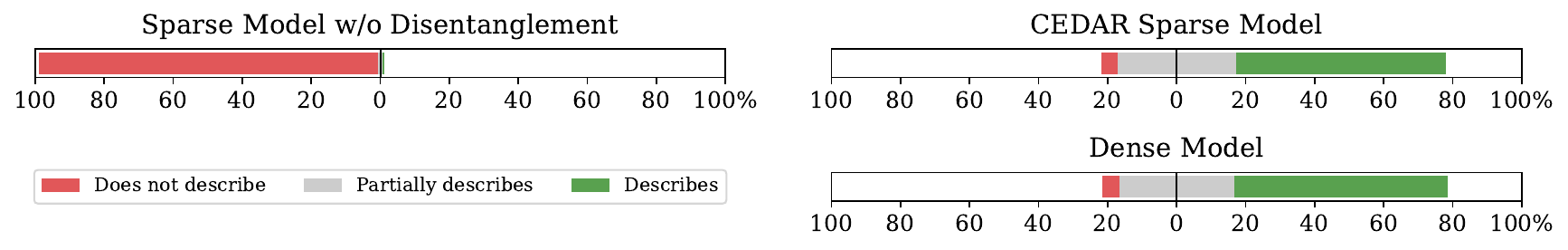}
    \caption{\our{} achieves ratings close to the dense model without sparsification and substantially outperforms the non-disentangled sparse approach.}
    \label{fig:method-scores-distribution}
\end{figure}

Figure~\ref{fig:method-scores-distribution} shows the distribution of ratings across methods. The variant without disentanglement, i.e., sparsification directly in the original embedding space, is dominated by low ratings, indicating that its captions rarely describe the image. In contrast, \our{} achieves a substantially higher proportion of top ratings and closely matches the dense baseline using full embeddings without sparsification. This indicates that learning the transformation $U$ is essential for preserving semantic information under sparse compression.

\section{Evaluating the compressed embeddings} \label{sec:bench}

In this section, we evaluate the quality of compressed embeddings without referring to their text interpretations. 

\paragraph{Experimental setup.}
\label{sec:exp_setup}

The experimental protocol closely follows the Matryoshka Sparse Autoencoder (MSAE)~\cite{zaigrajew2025interpreting} framework to ensure a fair and standardized comparison. CLIP~\cite{radford2021learning} with a ViT-L/14~\cite{dosovitskiy2021vit} image encoder is used as the backbone, and all methods operate on the class token representation. All methods (\our{} and 4 variants of SAE) are trained on ImageNet-1K ~\cite{deng2009imagenet} training set and evaluated on its validation split. Details of the experimental setup are included in Appendix \ref{sec:setup}.

To ensure a fair comparison between methods that differ in the dimensionality of the representation, we fix the average reconstruction error, FVU~\cite{gao2024scaling}, and evaluate the remaining metrics: the number of non-zero activations (K), information capacity of the representation (IC), cosine similarity between original and reconstructed representations (CS), cross-entropy of the linear probing on ImageNet validation labels (LP), and Centered Kernel Nearest Neighbor Alignment (CKNNA)~\cite{huh2024platonic}, see Appendix~\ref{sec:setup}.

\paragraph{How much information is needed for reconstruction?}

\begin{table}[h]
\caption{Comparison of the compressed embeddings under fixed reconstruction error levels.}
\label{tab:sae_comp}
\vspace*{0.1in}
\centering
\footnotesize
\begin{tabular}{llcccccc}
\toprule
$FVU$ & Model & $K \downarrow$ & $IC \downarrow$  & $CS \uparrow$ & $LP \downarrow$& $CKNNA \uparrow$ \\
\midrule
\multirow{5}{*}{0.25} & MSAE & 8.090 & 59.488 & 0.866 & 2.062 & 0.405 \\
&TopK & 7.001 & 52.206 & 0.866 & 2.068 & 0.444 \\
&ReLu & 27.305 & 171.395 & 0.870 & 2.456 & 0.339  \\
&BatchTopK & 4.207 & 32.939 & 0.866 & 1.946 & 0.416 \\
&\our{} & 11.549 & 57.381 & 0.865 & 2.087 & 0.425 \\
\midrule
\multirow{5}{*}{0.3}& MSAE & 5.322 & 40.798 & 0.837 & 2.690 & 0.354 \\
&TopK & 4.523 & 35.145 & 0.836 & 2.652 & 0.403 \\
&ReLu & 13.728 & 94.694 & 0.840 & 3.139 & 0.299 \\
&BatchTopK & 2.633 & 21.366 & 0.836 & 2.546 & 0.370 \\
&\our{} & 5.411 & 30.206 & 0.836 & 2.592 & 0.341 \\
\midrule
\multirow{5}{*}{0.35}& MSAE & 3.406 & 27.109 & 0.807 & 3.491 & 0.296 \\
& TopK & 2.786 & 22.461  & 0.805 & 3.413 & 0.352  \\
& ReLu & 6.828 & 50.911 & 0.810 & 3.978 & 0.247 \\
& BatchTopK & 1.591 & 13.286  & 0.804 & 3.327 & 0.323  \\
& \our{} & 2.690 & 16.272 & 0.806 & 3.351 & 0.247 \\
\bottomrule
\end{tabular}
\end{table}

Table~\ref{tab:sae_comp} analyzes the performance of disentanglement methods at three matched reconstruction error levels (FVU = 0.25, 0.3, 0.35), allowing us to study how much information is required to achieve a given reconstruction fidelity. By controlling FVU, we can directly compare methods in terms of the number of active features (K) and the corresponding information capacity (IC). Across all methods, lower reconstruction error consistently requires more active components and induces higher information capacity, indicating a clear trade-off between sparsity and reconstruction fidelity. Importantly, cosine similarity (CS) mirrors the behavior of FVU across settings, suggesting that it also reliably reflects reconstruction quality in this regime.

At the method level, BatchTopK~\cite{bussmann2024batchtopk} shows a generally favorable trade-off between sparsity and reconstruction quality, often requiring relatively few active components while maintaining competitive similarity scores. TopK~\cite{gao2024scaling} performs comparably, frequently matching BatchTopK~\cite{bussmann2024batchtopk} in reconstruction quality, though sometimes using slightly more features. In contrast, ReLU~\cite{bricken2023towards} operates in a less compressed regime overall, relying on substantially more active features and higher information capacity. Despite these differences, \our{} remains competitive across both sparsity and semantic preservation metrics, performing on par with the strongest baselines in most settings. Notably, in terms of information capacity (IC), \our{} ranks among the top-2 methods at higher sparsity regimes, indicating that it achieves comparable reconstruction fidelity while using less encoded information. An additional advantage of our approach is that it is inherently lossless in its default form, as it is based on an invertible transformation.




\paragraph{Are the reduced embeddings semantically meaningful?}

\begin{figure}[htbp]
    \centering

    \begin{subfigure}{0.45\textwidth}
        \centering
        \includegraphics[width=\textwidth]{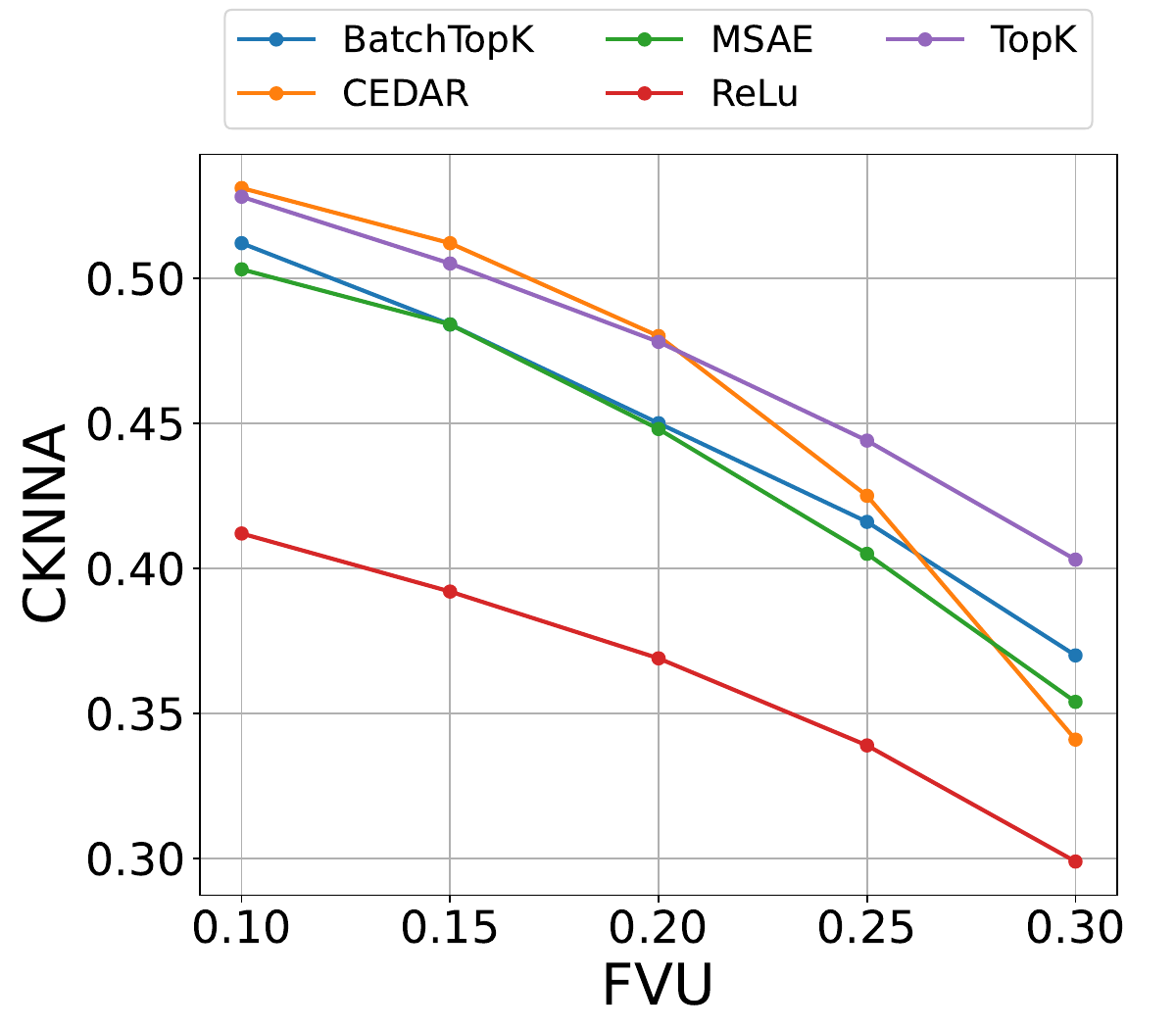}
    \end{subfigure}%
    \begin{subfigure}{0.45\textwidth}
        \centering
        \includegraphics[width=\textwidth]{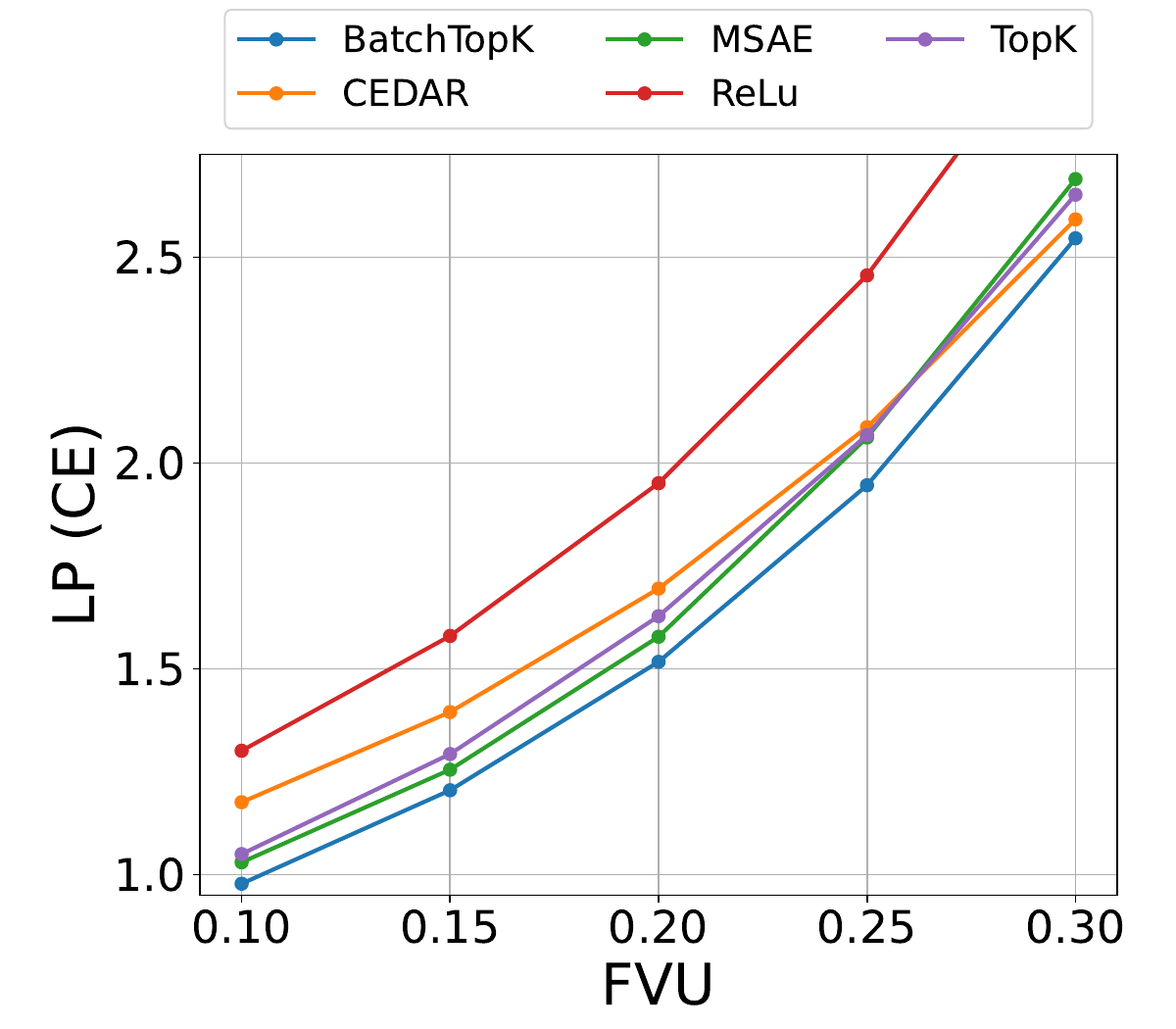}
    \end{subfigure}%
    \label{fig:class_quality}

    \caption{Semantic quality of reconstructed embeddings measured by CKNNA~\cite{huh2024platonic} (left) and linear probing cross-entropy (LP) (right) under varying FVU levels.}
    \label{fig:semantic_plots}
\end{figure}

Table~\ref{tab:sae_comp} also reports semantic quality metrics, including CKNNA, which captures structural alignment, and linear probing cross-entropy (LP), which reflects downstream task performance. These metrics allow us to evaluate how well the reduced embeddings preserve meaningful structure and class-relevant information. From the table, we observe noticeable variation across methods even at matched reconstruction levels, suggesting that similar reconstruction fidelity does not guarantee comparable preservation of semantic structure.

Figure~\ref{fig:semantic_plots} further illustrates how these metrics evolve as reconstruction error (FVU) decreases. As expected, increasing FVU leads to a gradual degradation in both metrics, indicating that reduced embeddings lose semantic information as reconstruction quality decreases. However, the rate of this degradation differs across methods. TopK and BatchTopK consistently maintain strong semantic alignment, with TopK achieving the highest CKNNA across most regimes, suggesting better preservation of neighborhood structure. In terms of downstream performance, BatchTopK and MSAE remain competitive, achieving lower cross-entropy compared to other methods at higher FVU levels. ReLU-SAE shows the weakest performance in both metrics, with significantly worse CKNNA and higher classification error, confirming that its denser representations do not translate into better semantic preservation. 

Our method follows trends similar to the strongest baselines, maintaining competitive performance in both CKNNA and linear probing across all FVU levels. While not always achieving the best scores, it preserves semantic structure and downstream performance at a comparable level, indicating that the reduced embeddings remain meaningfully aligned with the original representation space.








\input{sections/06_conclusions}
\medskip

\section*{Acknowlegdements}
The work of Jacek Tabor and Łukasz Struski was supported by the National Science Centre, Poland, grants no. 2023/49/B/ST6/01137. The research of Patryk Marszałek and Marek \'Smieja was supported by the National Science Centre (Poland), grant no. 2023/50/E/ST6/00169. Some experiments were performed on servers purchased with funds from the flagship project entitled ``Artificial Intelligence Computing Center Core Facility'' from the DigiWorld Priority Research Area within the Excellence Initiative -- Research University program at Jagiellonian University in Kraków.

The contribution of Piotr Kubaty, Patryk Marszałek, and Adam Wróbel to this research was conducted at the Faculty of Mathematics and Computer Science, and the Doctoral School of Exact and Natural Sciences of the Jagiellonian University.

\bibliographystyle{unsrt}
\bibliography{ref}


\newpage
\appendix

\section{Formulation of the Training Curriculum}
\label{app:training_curriculum}
We consider a homotopy-style curriculum:
\[
k(t) = D + \frac{t}{\tau}(k_{\mathrm{target}} - D), \quad t \le \tau,
\]

which gradually alters the objective from identity reconstruction ($t=0$, $k(t)=D$) to a sparse projection ($t=\tau$, $k(t) = k_{\mathrm{target}}$).

This can be interpreted as solving a sequence of optimization problems:
\[
\min_{U} \mathbb{E}\left[\| \mathbf{z} - \hat{\mathbf{z}}^{(k)} \|_1 \right],
\quad k = D \rightarrow k_{\mathrm{target}},
\]
where each stage initializes the next.
Once the target sparsity regime is reached, we introduce stochasticity by sampling:
\[
k \sim \mathcal{U}(\{1, \ldots, k_{\mathrm{max}}\}),
\]
which prevents overfitting to a fixed support size and encourages robustness of the learned basis across different sparsity levels. We set $k_{\mathrm{target}}$ to 10, as bigger values would harm interpretability of the sparse model, and $k_\mathrm{max}$ to $2 k_{\mathrm{target}} -1$ to maintain expected value of $k$ at $k_{\mathrm{target}}$. In practice, this leads to more stable support sets $\mathcal{S}$ and reduces sensitivity to the exact choice of $k$ at inference time.

\section{Details of the experimental setup} \label{sec:setup}

\emph{Datasets and baselines.} We compare against MSAE~\cite{zaigrajew2025interpreting} (reverse weighting)~\cite{zaigrajew2025interpreting}, TopK-SAE (k = 64)~\cite{gao2024scaling}, ReLU-SAE~\cite{bricken2023towards} ($\lambda$ = 0.01)~\cite{bricken2023towards}, and BatchTopK-SAE (k = 32)~\cite{bussmann2024batchtopk}, all trained on ImageNet-1K training set and evaluated on its validation split. Similarly to \our{}, SAE models are standardized using the mean of CLIP~\cite{radford2021learning} representations computed on the training set, with $\sqrt{n}$ used as a scaling factor. Our method uses the default 768-dimensional CLIP~\cite{radford2021learning} embedding, while SAE baselines employ an 8$\times$ expansion to 6144 dimensions. All baselines are implemented using the official MSAE~\cite{zaigrajew2025interpreting} codebase with default hyperparameters, except for the specific settings explicitly stated above.

\emph{Metrics.} In the experiment sections, we evaluate models using a range of complementary metrics. Fraction of Variance Unexplained (FVU), also known as normalized MSE, measures reconstruction quality by relating the reconstruction error to the variance of mean-centered inputs. We denote by $K$ the number of non-zero activations in the sparse representation. Information capacity (IC) quantifies the amount of information that can be encoded using the $k$ of $n$ coordinates, which is defined as $\log \binom{n}{k}$, where $n$ is the representation dimensionality (768 for \our{}, 6144 for SAE baselines) and $k$ is the number of non-zero neurons activations. The metric is computed per image sample and then averaged over the dataset. Alignment between original and reconstructed representations is captured using cosine similarity (CS). Semantic preservation is further evaluated via linear probing. Following the MSAE~\cite{zaigrajew2025interpreting} pipeline, a linear classifier is trained on original representations with AdamW optimizer and a ReduceLROnPlateau scheduler. Performance is then measured on reconstructed embeddings using cross-entropy on ImageNet validation labels. Finally, Centered Kernel Nearest Neighbor Alignment (CKNNA)~\cite{huh2024platonic} measures structural consistency by comparing mutual nearest neighbor relationships in the original and reconstructed feature spaces.

To ensure a fair comparison between methods that differ in the dimensionality of the sparse representation, we fix the average reconstruction error (FVU) and evaluate the remaining metrics. For each method, this target is matched by controlling the activation level in the sparse space, where activations below a threshold are set to zero. The threshold is selected independently for each method using binary search to achieve the desired FVU level.

\section{Additional Qualitative Results of \our{}}
\label{app:additional_results}

\begin{figure}[h!]
\centering

\begin{subfigure}{0.49\textwidth}
\includegraphics[width=\linewidth]{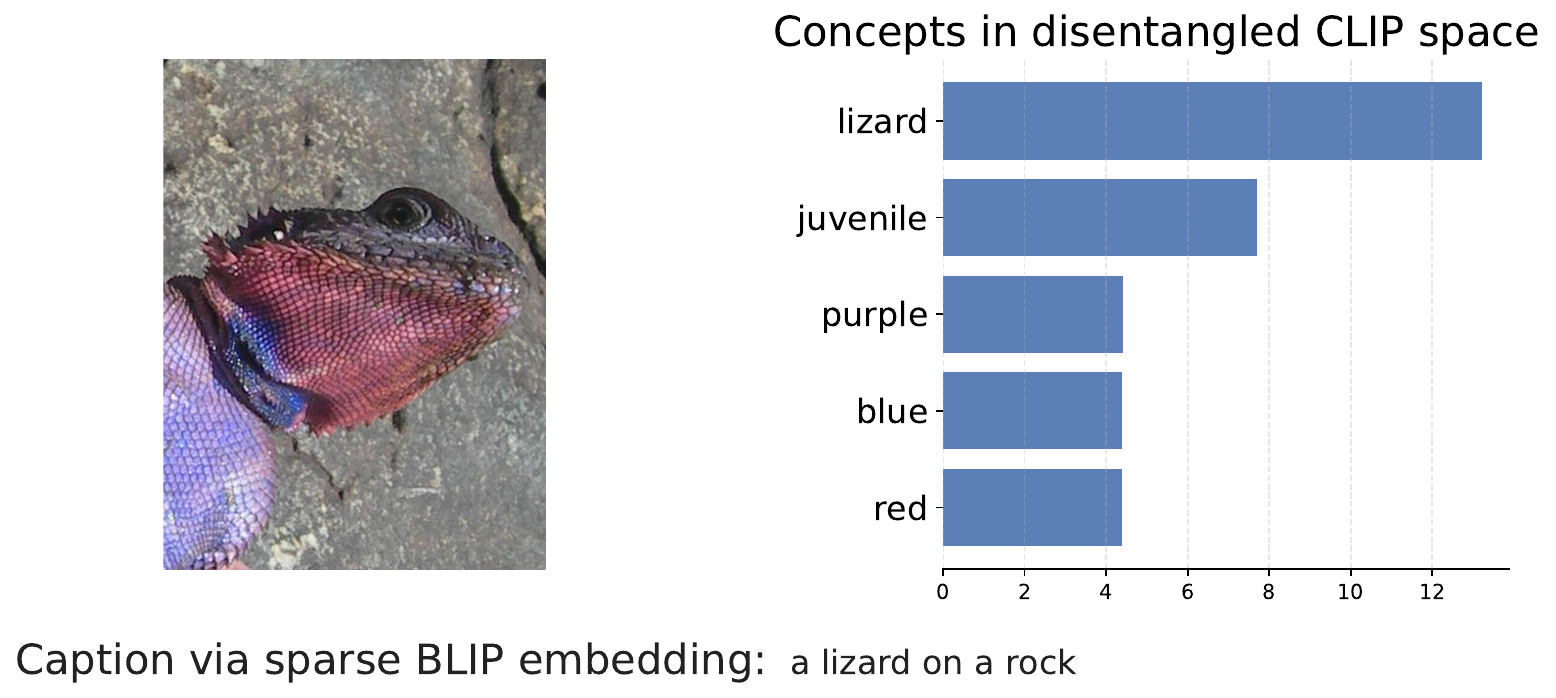}
\end{subfigure}
\hfill
\begin{subfigure}{0.49\textwidth}
\includegraphics[width=\linewidth]{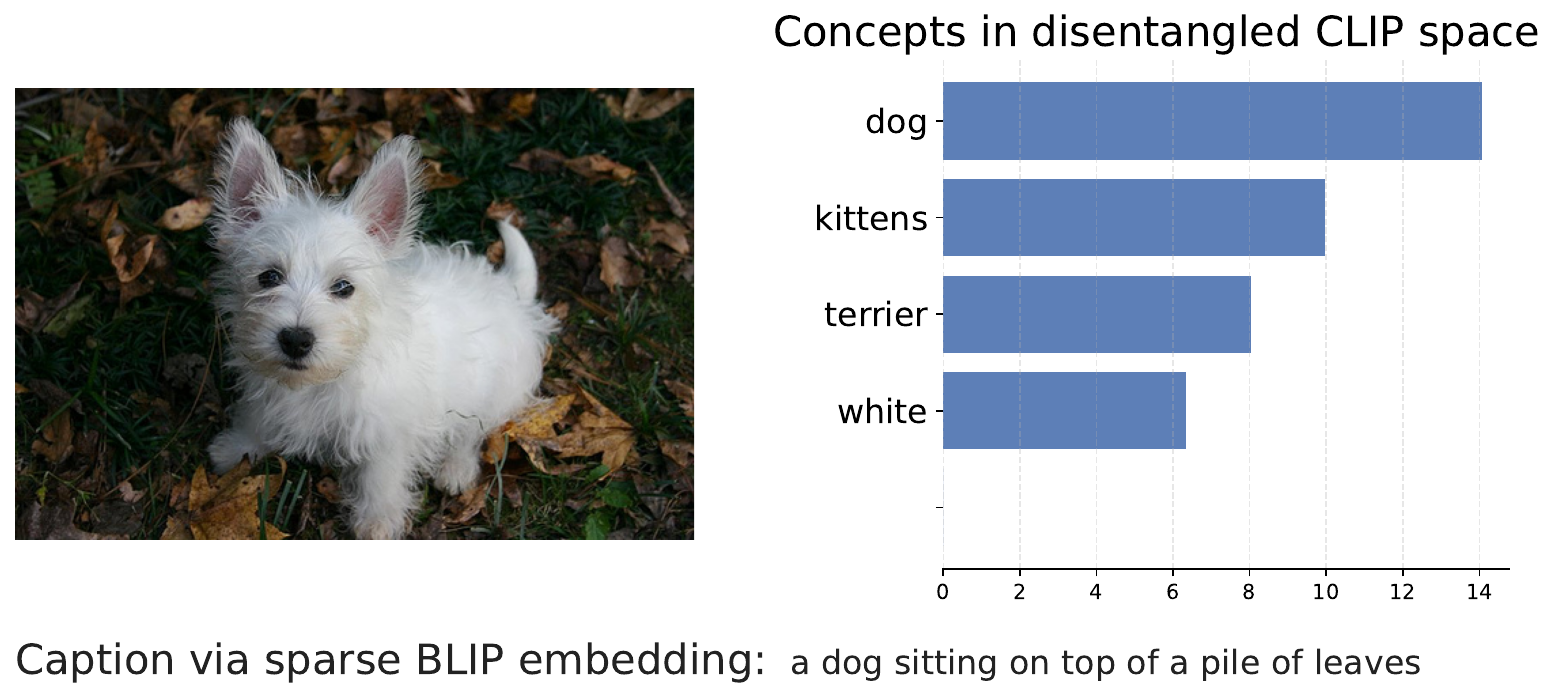}
\end{subfigure}

\vspace{0.5cm}

\begin{subfigure}{0.49\textwidth}
\includegraphics[width=\linewidth]{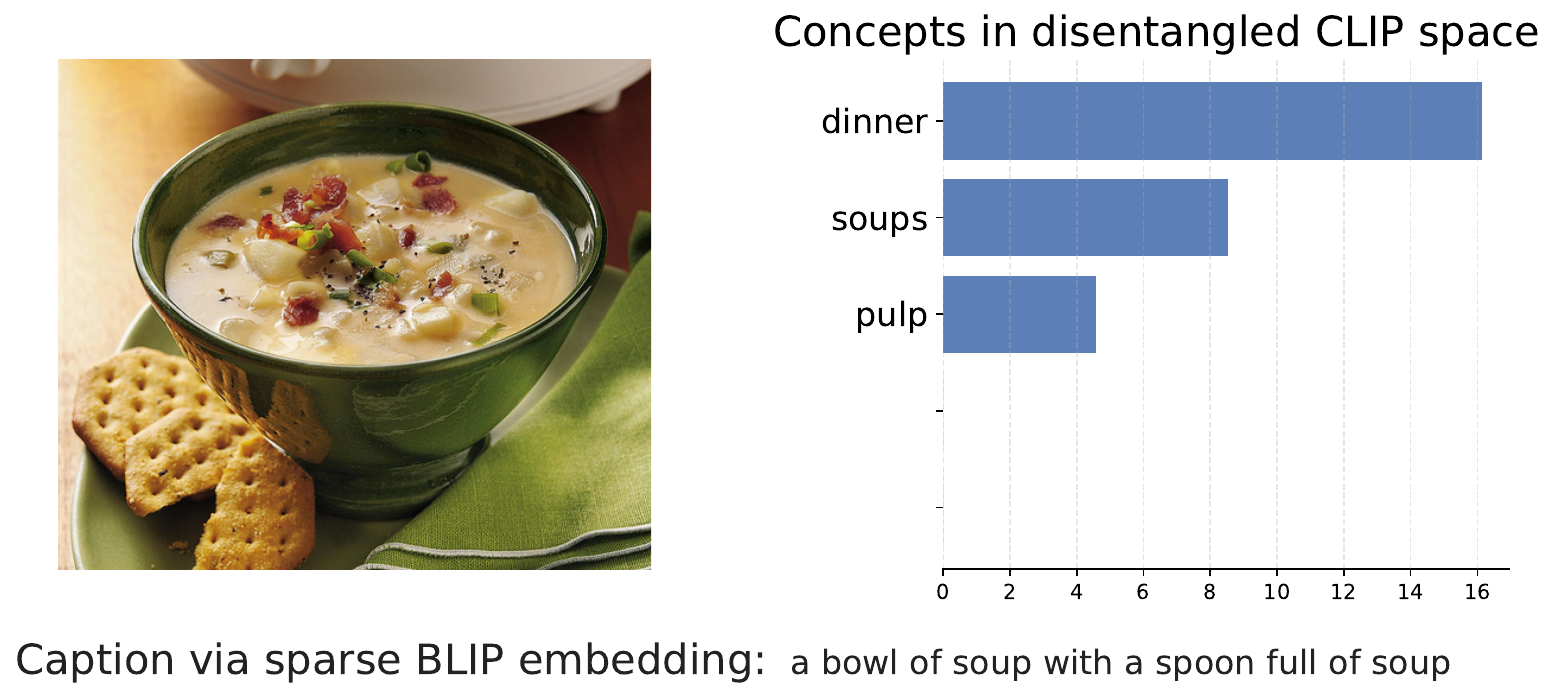}
\end{subfigure}
\hfill
\begin{subfigure}{0.49\textwidth}
\includegraphics[width=\linewidth]{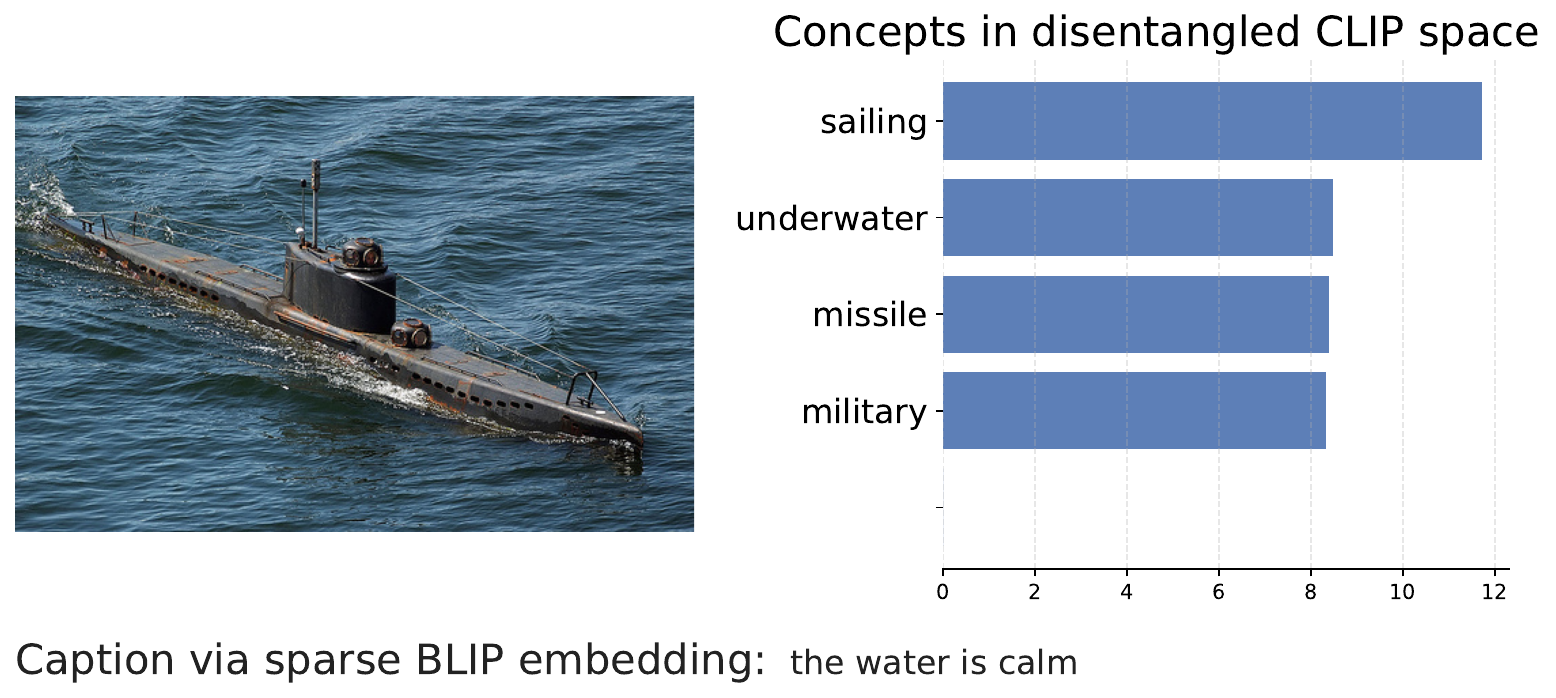}
\end{subfigure}

\vspace{0.5cm}

\begin{subfigure}{0.49\textwidth}
\includegraphics[width=\linewidth]{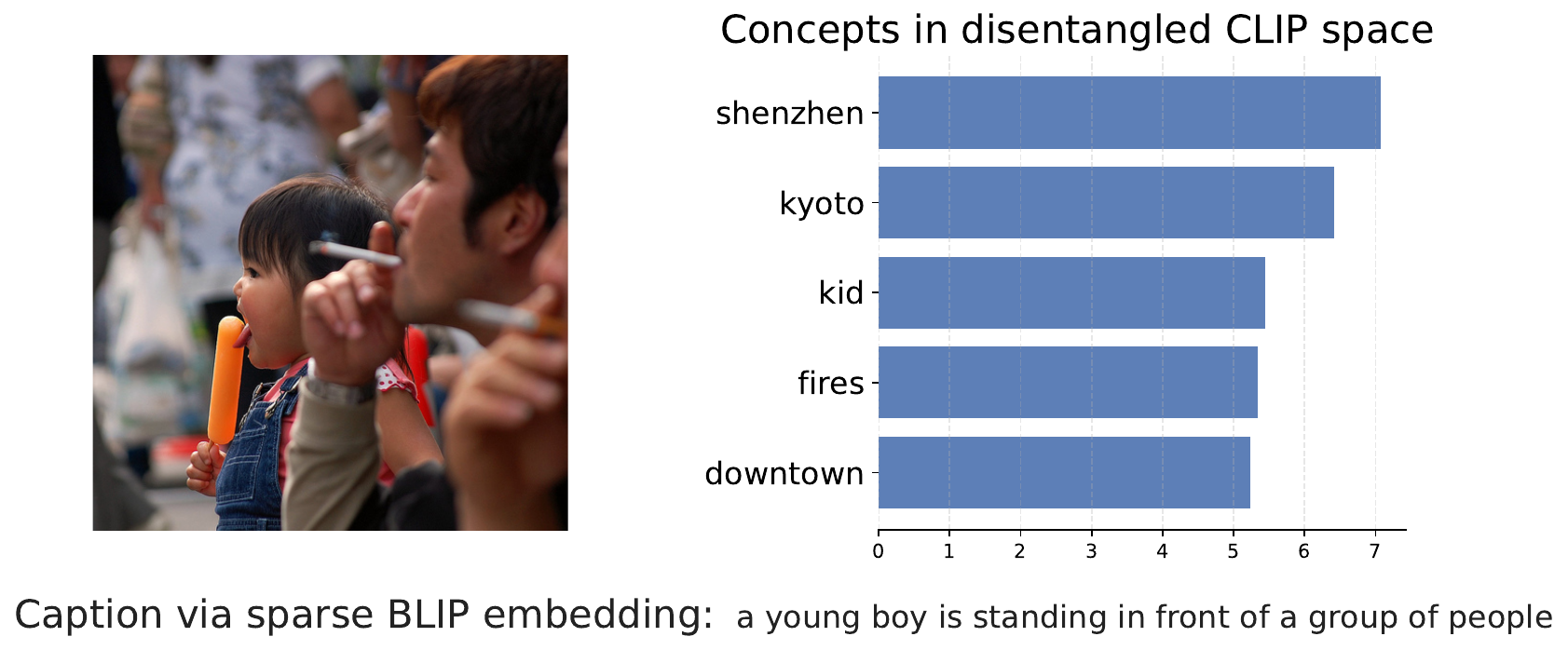}
\end{subfigure}
\hfill
\begin{subfigure}{0.49\textwidth}
\includegraphics[width=\linewidth]{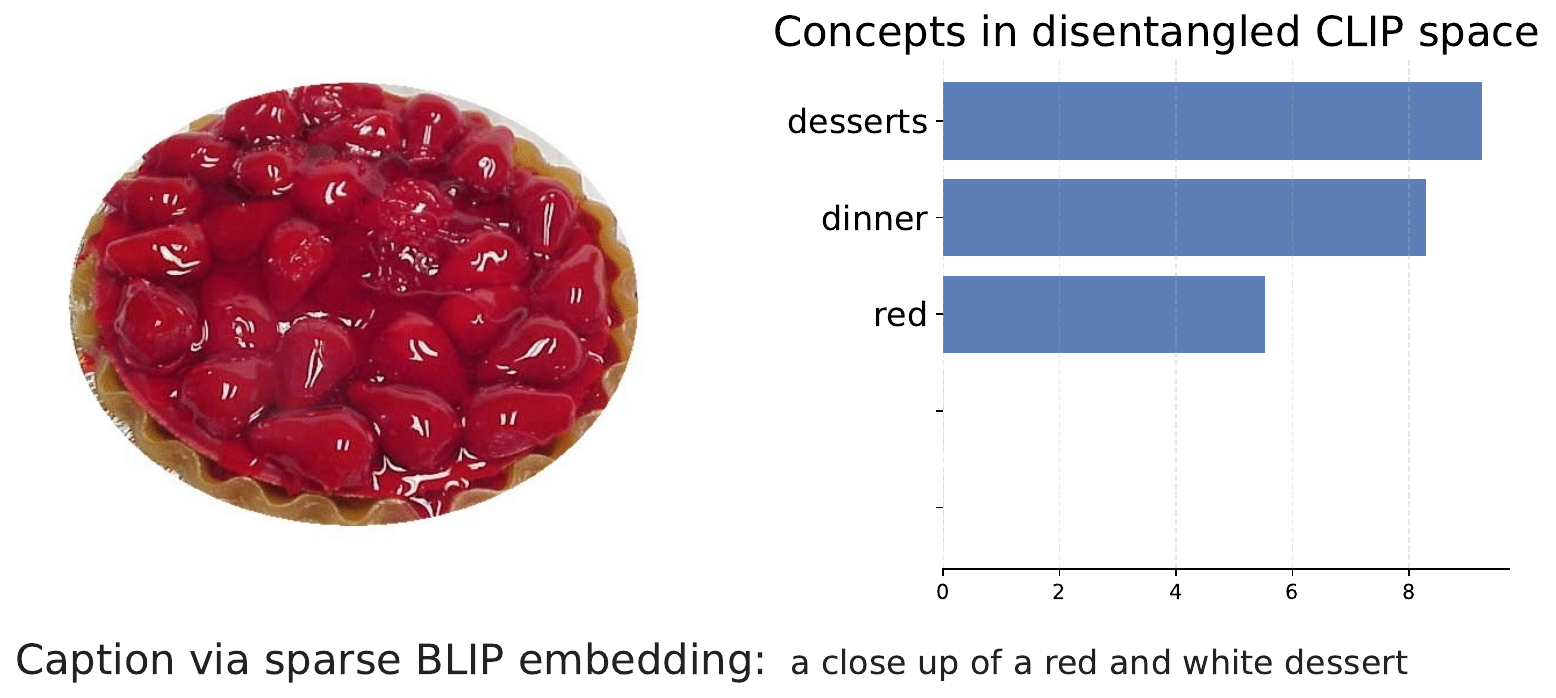}
\end{subfigure}

\caption{Further qualitative results illustrating \our{}: human-readable concepts derived from the disentangled CLIP~\cite{radford2021learning} space alongside captions produced from sparse BLIP~\cite{li2022blip} embeddings.}
\label{fig:additional_text_caption}
\end{figure}

We first present more visual examples of CEDAR interpretations, extending the analysis from Section~\ref{sec:ill_examples}. Figure~\ref{fig:additional_text_caption} provides further evidence of the model's ability to maintain semantic consistency across diverse image categories while enforcing sparsity in the BLIP~\cite{li2022blip} embedding space and ensuring clear concept alignment in the disentangled CLIP~\cite{radford2021learning} space.

\begin{figure}
    \centering
    \includegraphics[width=\linewidth]{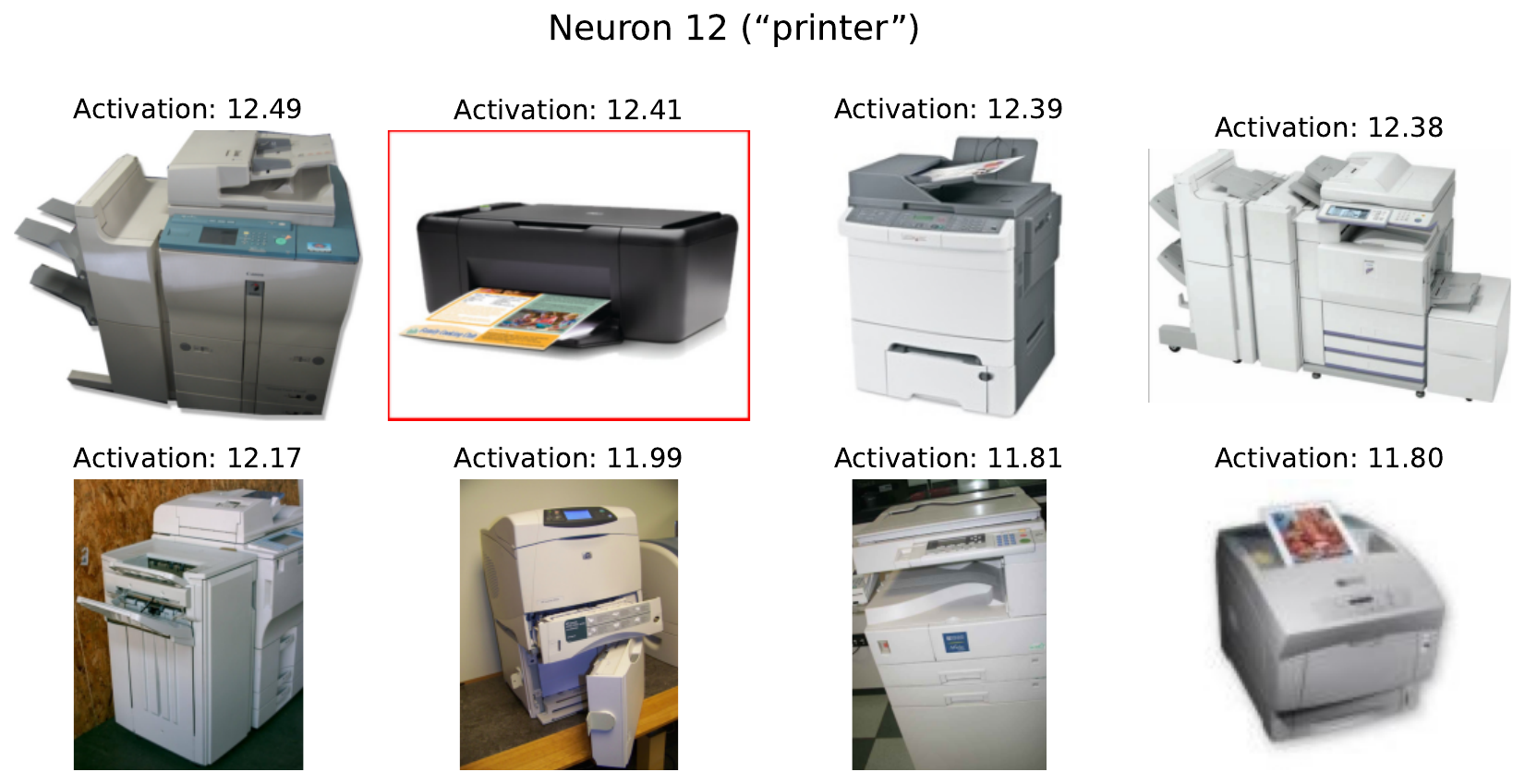}
    \caption{Top-activating samples for neuron 12 in the disentangled representation, aligned with the concept ``printer''.}
    \label{fig:concept_printer}
\end{figure}

\begin{figure}
    \centering
    \includegraphics[width=\linewidth]{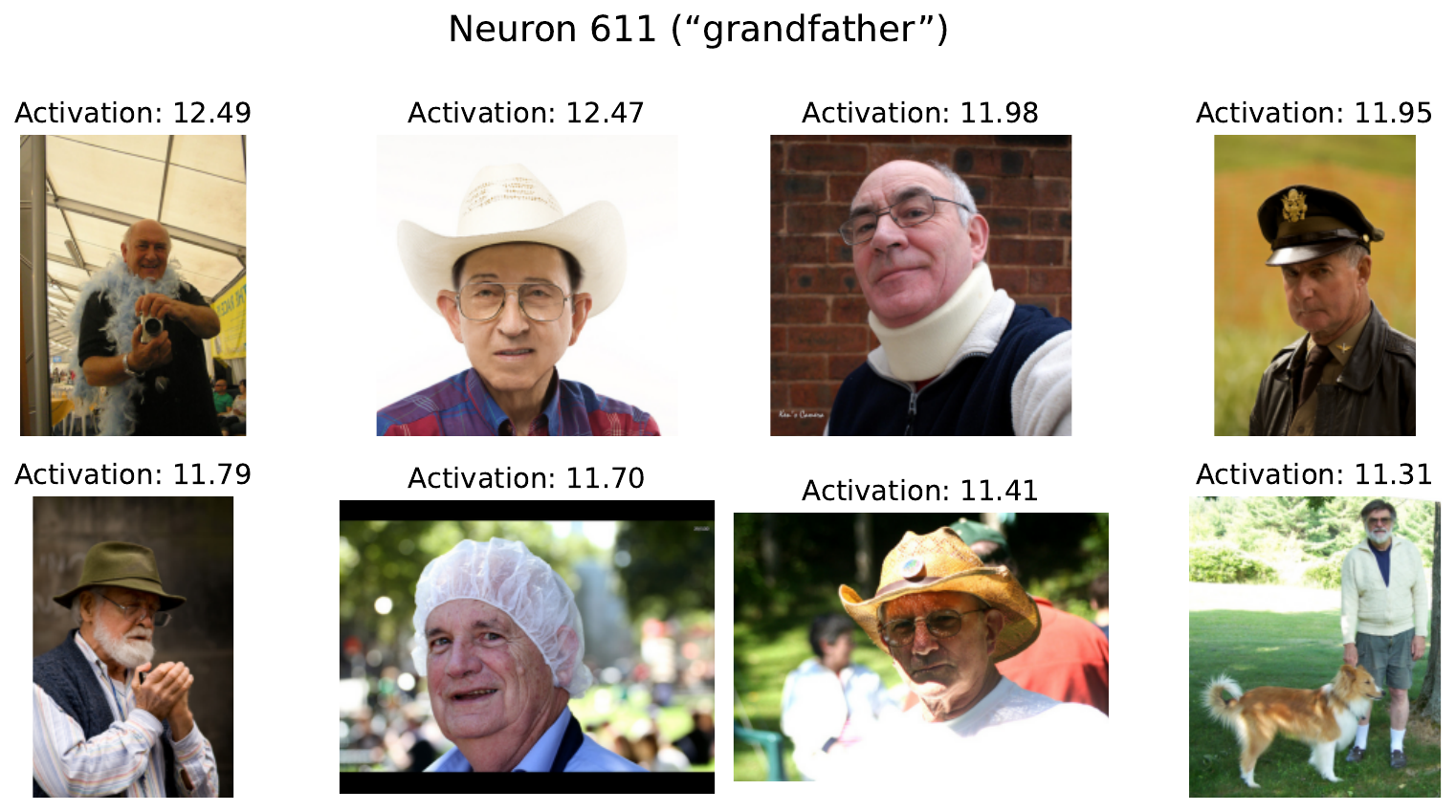}
    \caption{Examples eliciting the strongest responses from neuron 611 in the disentangled space, corresponding to the concept ``grandfather''.}
    \label{fig:concept_grandfather}
\end{figure}

\begin{figure}
    \centering
    \includegraphics[width=\linewidth]{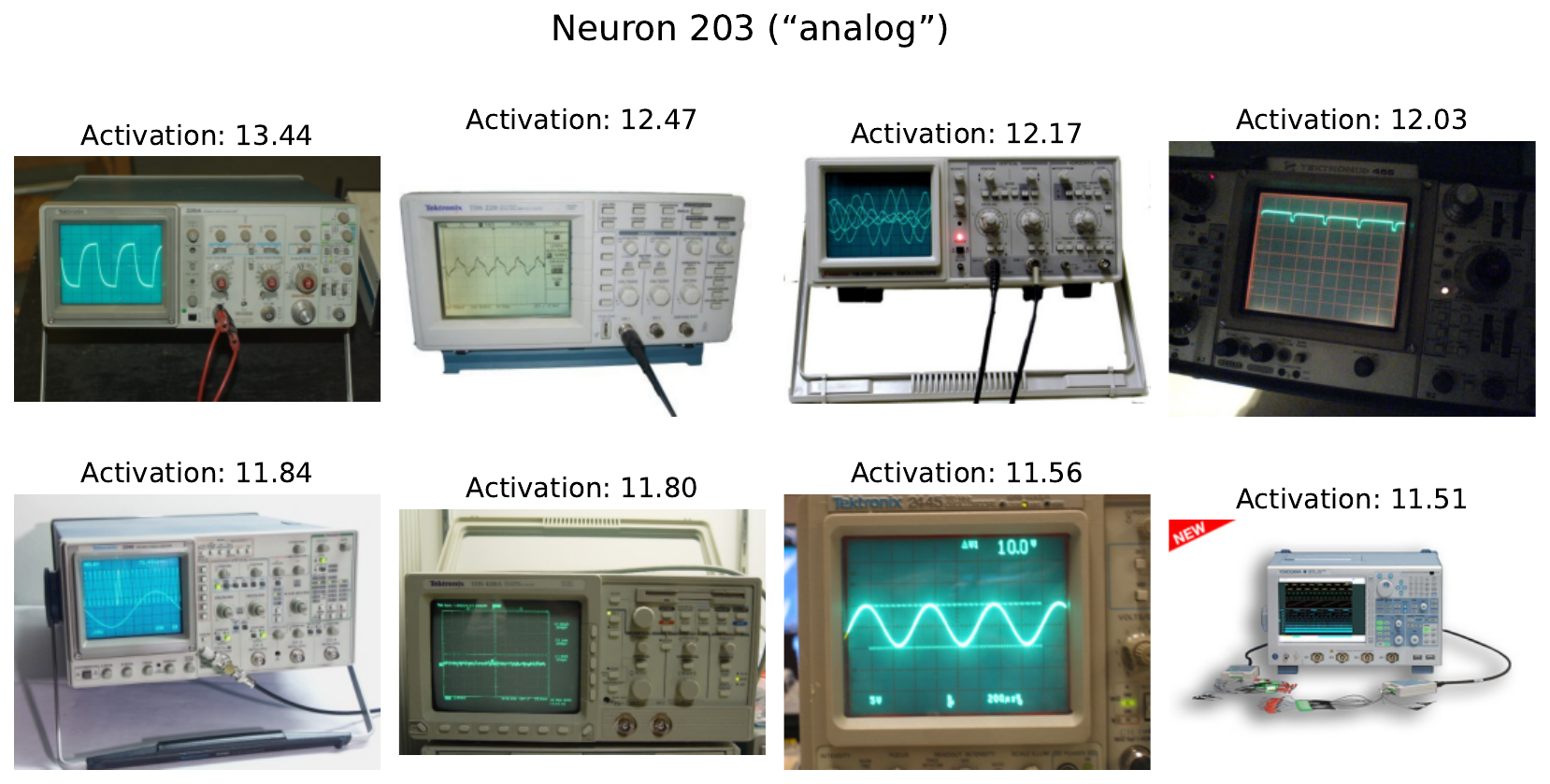}
    \caption{Samples that maximally activate neuron 203 within the disentangled embedding, reflecting the concept ``analog''.}
    \label{fig:concept_analog}
\end{figure}

Furthermore, we visualize the semantic purity of specific neurons in the disentangled space. Figure~\ref{fig:concept_printer} illustrates top activations for the "printer" concept, while Figure~\ref{fig:concept_grandfather} and Figure~\ref{fig:concept_analog} demonstrate the model's capacity to isolate complex social roles and specific technological styles, respectively. Collectively, these results reinforce that the disentangled dimensions represent stable and human-interpretable visual primitives.

\FloatBarrier

\section{User Study}
\subsection{Detailed Protocol}
\label{app:user-study-details}

User studies were conducted using Google Forms. Studies 1, 2, and 3 involved 23, 22, and 21 participants, respectively.

\paragraph{Study 1: Method Preference}
The study consisted of 8 independent comparison tasks. In each task, participants were shown a single image along with two concept-based explanations, labeled A and B. Each explanation was represented as a list of five concepts.

The assignment of methods (\our{} and MSAE~\cite{zaigrajew2025interpreting}) to labels A and B was independently randomized for each task to mitigate ordering bias. Participants were instructed to rate their preference between the two explanations using a 5-point scale: \textit{Strongly prefer A}, \textit{Slightly prefer A}, \textit{Equally prefer A and B}, \textit{Slightly prefer B}, and \textit{Strongly prefer B}.

\paragraph{Study 2: Concept Selection}
The study consisted of 8 independent tasks. In each task, participants were presented with a single image and a set of six concepts. These concepts were drawn from two methods: three from \our{} and three from MSAE~\cite{zaigrajew2025interpreting}.

All six concepts were displayed simultaneously and their order was randomized for each task to avoid method-specific grouping effects. Participants were instructed to select all concepts they considered relevant to the image, with no restriction on the number of selections.

\paragraph{Study 3: Description Quality}
The study consisted of 10 independent tasks. In each task, participants were shown a single image along with three textual descriptions generated by different models: the dense model, our method, and a sparse model without a disentanglement mechanism. The order of the descriptions was independently randomized for each task.

Participants were asked to evaluate how well each description reflects the image content, focusing on the overall impression rather than fine-grained details. Each description was independently rated using a 3-point ordinal scale: \textit{1 -- Does not describe the image}, \textit{2 -- Partially describes the image}, and \textit{3 -- Describes the image}. Participants were allowed to assign the same rating to multiple descriptions.

\subsection{Instructions and Questionnaire}
\label{app:user-study-examples}
This section provides examples of the instructions, task descriptions, and survey questions used in each study.
\begin{figure}[h!tbp]
    \centering
    \begin{subfigure}{0.49\textwidth}
        \centering
        \includegraphics[width=\linewidth]{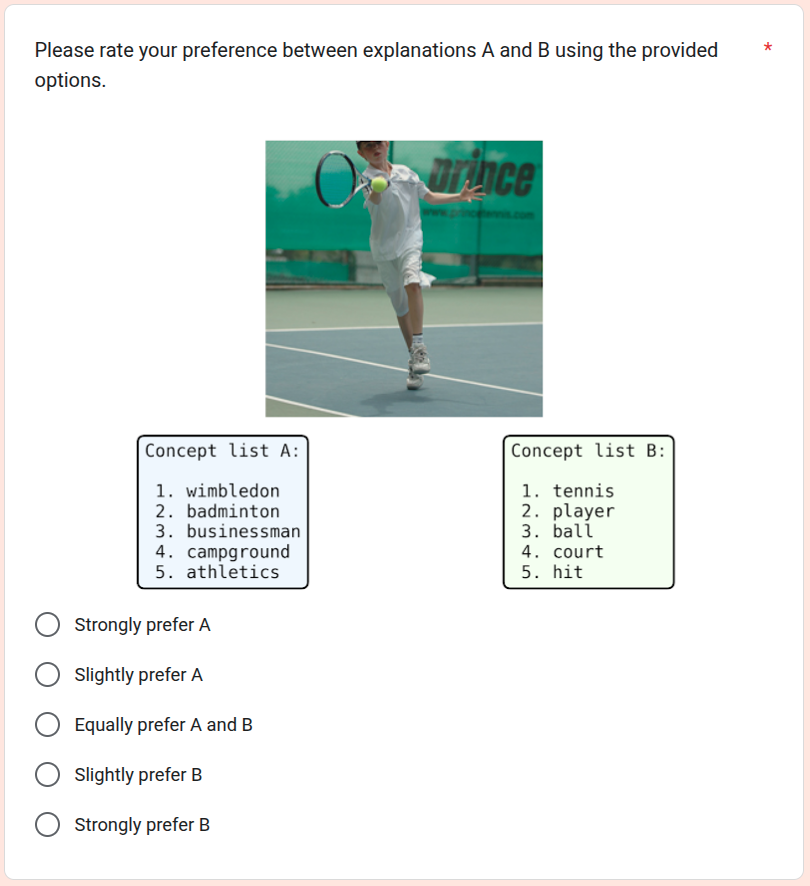}
    \end{subfigure}
    \hfill
    \begin{subfigure}{0.49\textwidth}
        \centering
        \includegraphics[width=\linewidth]{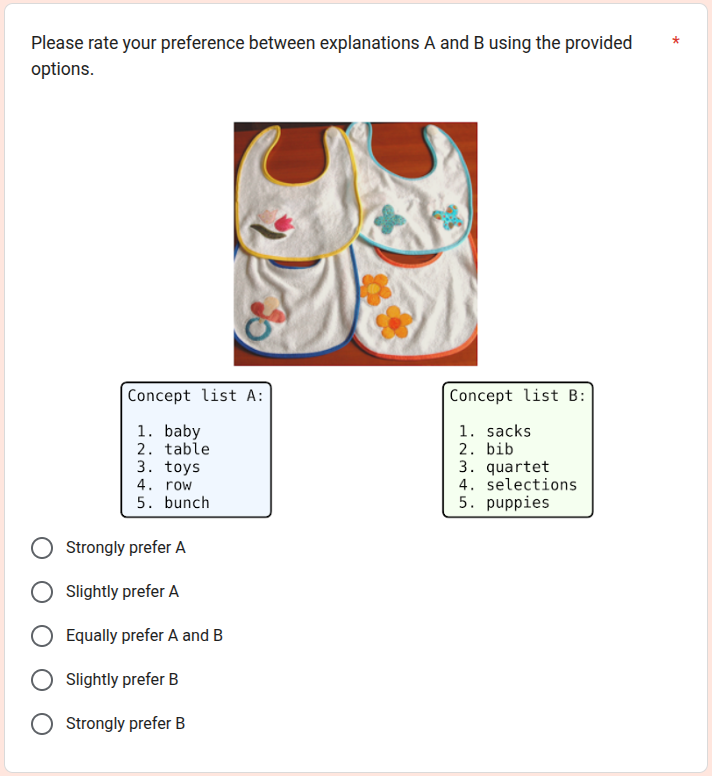}
    \end{subfigure}
    \caption{User interface screenshots for two example tasks in Study 1. In the screenshot on the left, \our{} explanation is shown as Concept list B, wheras in the one on the right, \our{} explanation is shown as Concept list A}
    \label{fig:user_study_screen_task_1}
\end{figure}

\begin{figure}[h!tbp]
    \centering
    \begin{subfigure}{0.49\textwidth}
        \centering
        \includegraphics[width=\linewidth]{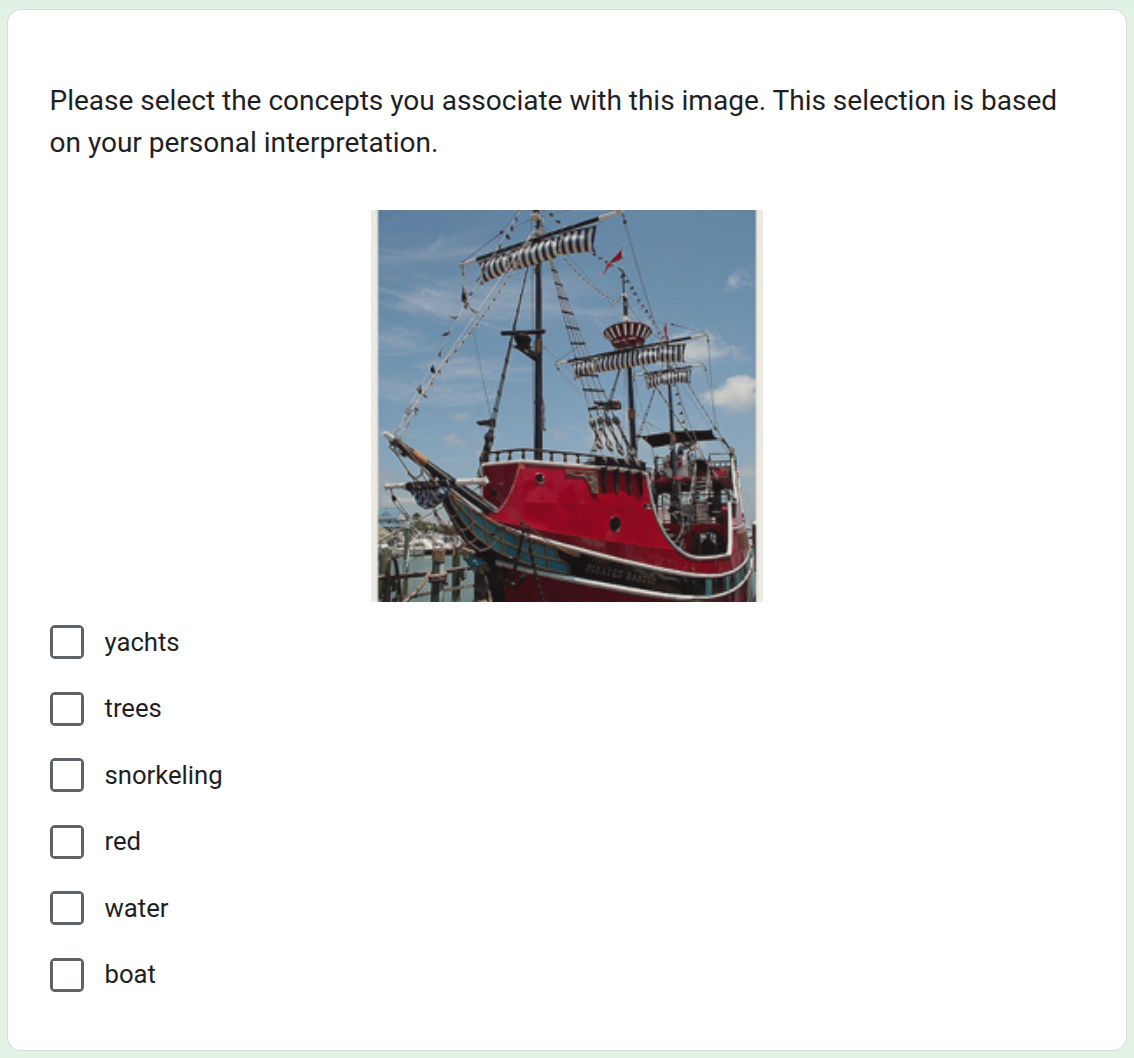}
    \end{subfigure}
    \hfill
    \begin{subfigure}{0.49\textwidth}
        \centering
        \includegraphics[width=\linewidth]{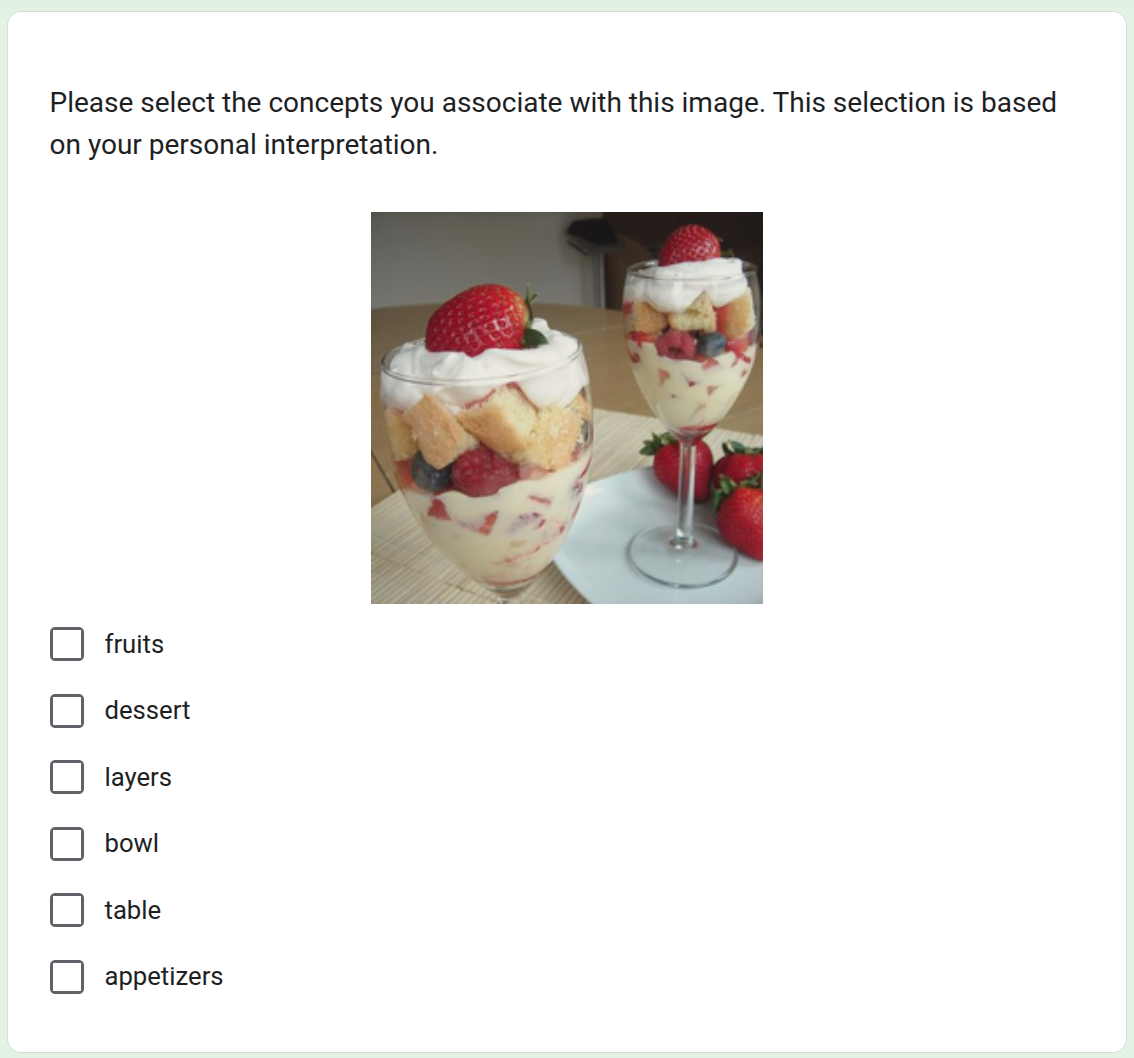}
    \end{subfigure}
    \caption{User interface screenshots for two example tasks in Study 2. In the screenshot on the left, \our{} concepts are: trees, water, boat, and in the one on the right, \our{} concepts are: dessert, bowl, table}
    \label{fig:user_study_screen_task_2}
\end{figure}

\begin{figure}[h!tbp]
    \centering
    \begin{subfigure}{0.49\textwidth}
        \centering
        \includegraphics[width=\linewidth]{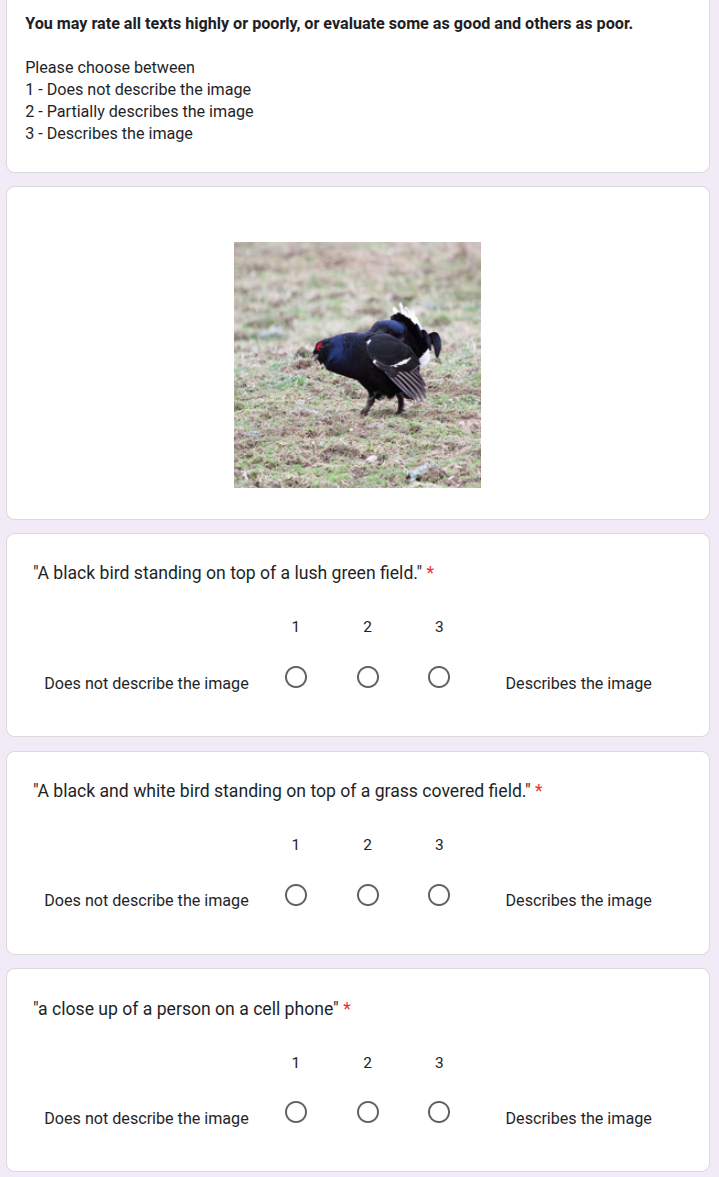}
    \end{subfigure}
    \hfill
    \begin{subfigure}{0.49\textwidth}
        \centering
        \includegraphics[width=\linewidth]{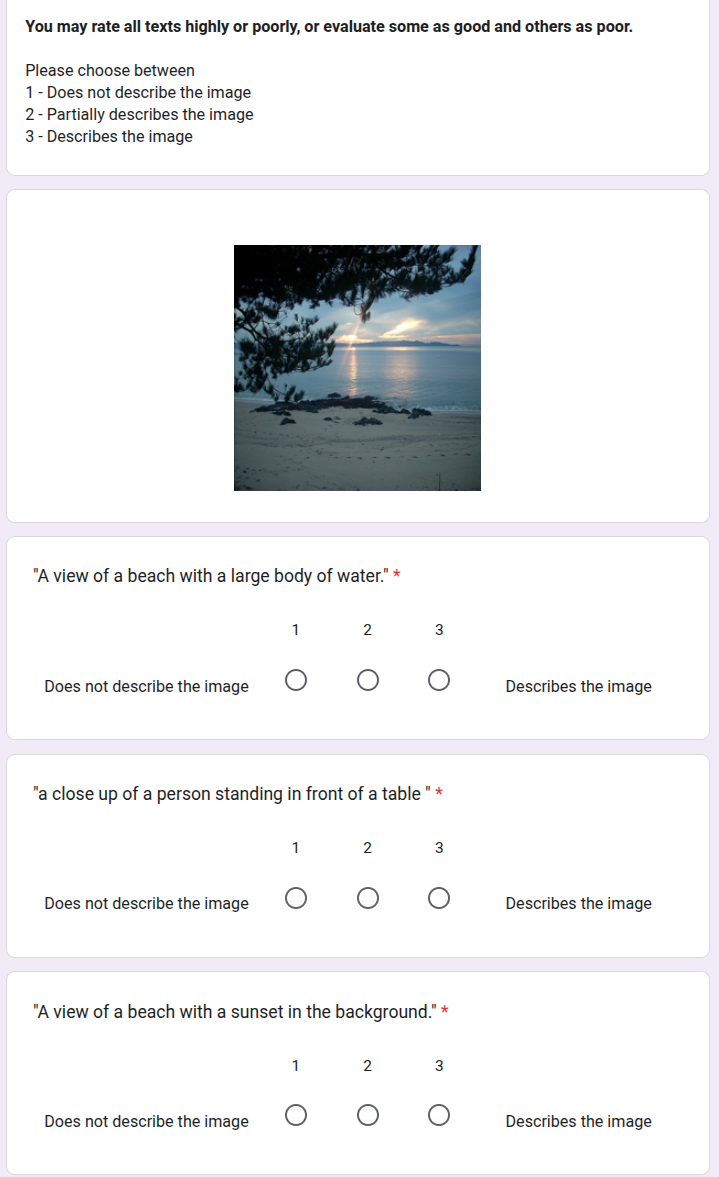}
    \end{subfigure}
    \caption{User interface screenshots for two example tasks in Study 3. Descriptions generated by \our{} are "A black bird standing on top of a lush green field." and "A view of a beach with a large body of water."}
    \label{fig:user_study_screen_task_3}
\end{figure}

\section{Broader Impacts}
\label{app:impacts}
CEDAR is designed to improve the interpretability of pretrained vision-language representations by learning sparse and semantically structured reparameterizations of embedding spaces. By exposing which latent coordinates are active for a given input and aligning these coordinates with textual concepts, the method provides a more transparent view into the internal organization of large pretrained models. Since CEDAR operates post-hoc on frozen representations, it can also be applied without modifying the original model or requiring costly retraining. At the same time, the semantic concepts associated with individual coordinates are obtained through similarity to text embeddings and therefore do not constitute causal or guaranteed explanations of model behavior. Sparse decompositions may also inherit biases and spurious correlations present in the underlying pretrained representations. Overall, we believe that methods such as CEDAR can contribute toward more transparent and analyzable multimodal systems, while emphasizing that interpretability outputs require careful human interpretation and validation.


\end{document}

%% file: sections/00_abstract.tex
\begin{abstract}
Vision-language models learn powerful multimodal embeddings, yet their internal semantics remain opaque. While sparse autoencoders (SAEs) can extract interpretable features, they rely on expanding the representation dimension, which compromises the original geometry and introduces redundancy. We introduce \our{} (Conceptual Embedding Disentanglement via Adaptive Rotation), a post-hoc method that reveals the compositional structure of pretrained embeddings without increasing dimensionality. By learning an invertible transformation with a top-$k$ sparsity bottleneck, \our{} concentrates semantic information into axis-aligned disentangled coordinates. In CLIP-like architecture, individual coordinates can be interpreted with textual concepts, while for generative models such as BLIP, they can be decoded into natural language descriptions. 
Experiments demonstrate that \our{} achieves a competitive reconstruction-sparsity trade-off while producing explanations that are more interpretable and better aligned with human perception. Our results suggest that the apparent entanglement in vision-language representations can be resolved through a suitable change of basis, eliminating the need for overcomplete expansions.
\end{abstract}

%% file: sections/02_related_works.tex
\begin{figure}
    \centering
    \includegraphics[width=\linewidth]{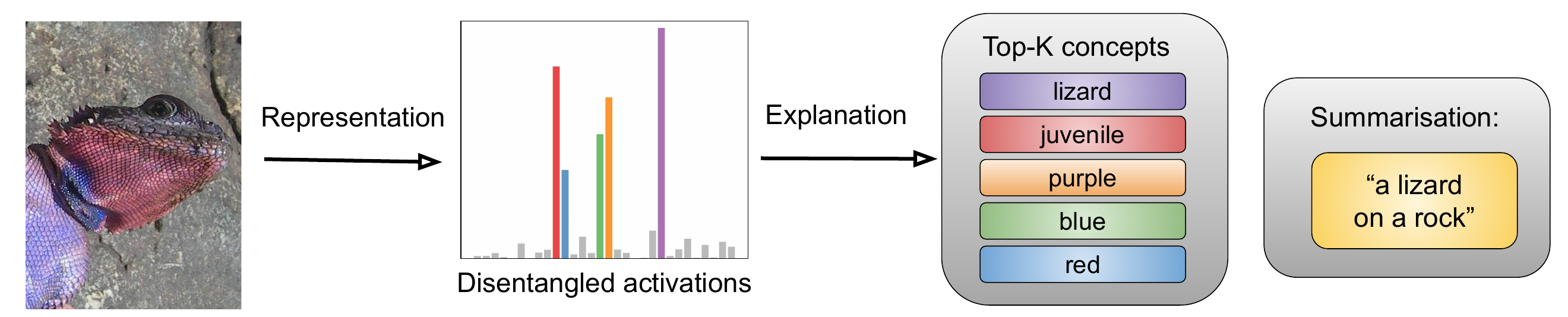}
    \caption{Overview of the \our{} pipeline. We disentangle representations in vision--language models into sparse components. Depending on the model architecture, we then extract semantic concepts associated with individual neurons or generate sparse textual descriptions.}
    \label{fig:placeholder}
\end{figure}

\section{Related works}

Vision-language models such as CLIP~\cite{radford2021learning} learn powerful multimodal embeddings, but their internal structure remains difficult to interpret. Most existing approaches focus on \emph{local explanations}, such as attribution maps or saliency methods~\cite{sundararajan2017axiomatic, bach2015pixel, abnar2020quantifying, chefer2021transformer, wrobel2026dave}, which identify input regions relevant to a specific prediction. While useful, these techniques provide limited insight into how semantic information is globally organized within the embedding space.

Concept-based methods~\cite{koh2020concept, ghorbani2019towards, kim2018interpretability} aim to connect model representations with human-understandable concepts. Approaches such as \emph{Testing with Concept Activation Vectors} (TCAV)~\cite{kim2018interpretability} and linear probing identify directions in representation space corresponding to predefined concepts, often leveraging text embeddings in vision-language models. More recent work explores automated concept discovery and alignment without manual supervision. However, these approaches operate within the original embedding geometry and implicitly assume that meaningful concepts are already linearly separable. As a result, they do not attempt to restructure the representation to produce compact, sparse, or disentangled concept representations.

A prominent line of work in mechanistic interpretability uses sparse autoencoders (SAEs)~\cite{bricken2023monosemantic, cunningham2024scaling, gao2025topk, rajamanoharan2024gated, zaigrajew2025interpreting} to uncover interpretable features in pretrained representations. These methods learn overcomplete dictionaries in which individual units correspond to approximately monosemantic features, addressing the problem of feature superposition in neural networks. By expanding the dimensionality and enforcing sparsity, SAEs enable the recovery of interpretable structure across a range of domains, including language and vision models. However, this approach relies on the assumption that disentangled features cannot be expressed in the original basis and require higher-dimensional representations. Consequently, the learned representations are not unique, and the expansion alters the geometry of the original embedding space.

Recent work has explored improving interpretability by reorganizing pretrained embeddings without modifying the underlying model. Methods such as EPIC~\cite{borycki2026epic,struski2024infodisent} and PluGeN~\cite{wolczyk2022plugen,proszewska2024multi,suwala2024face} decompose representations into structured components or align them with semantic attributes. While these approaches demonstrate that reparameterization can improve interpretability, they do not explicitly enforce sparsity, axis alignment, or invertibility, and therefore do not yield a global coordinate system in which semantic factors are directly readable.

Producing compact and information-preserving explanations remains a central challenge. Bottleneck-based approaches restrict representations to a small number of variables, often trading off interpretability against reconstruction fidelity~\cite{srivastava2024vlg,yang2023language,rao2024discover}. In contrast, we maintain the original dimensionality of the embedding space and instead enforce sparsity in a learned basis. This allows us to study highly compressed, low-activation regimes while preserving information through invertibility.

In contrast to prior work, we propose to learn a global, invertible reparameterization of the embedding space that induces sparsity and axis alignment directly in the original dimensionality. This formulation tests the hypothesis that much of the apparent entanglement in modern embeddings arises from basis misalignment rather than insufficient dimensionality. By combining sparsity, invertibility, and alignment with language, our method yields compact, globally consistent, and human-interpretable representations without expanding the embedding space.

%% file: sections/06_conclusions.tex
\section{Conclusions}
\label{sec:conclusions}
We introduced \our{}, a post-hoc method for interpreting vision-language embeddings through sparse, disentangled, and text-aligned representations within the original embedding space. In contrast to sparse autoencoder-based approaches, which rely on overcomplete expansions, our method learns an invertible reparameterization that induces sparsity without increasing dimensionality.
\our{} achieves a reconstruction–sparsity trade-off comparable to existing methods while preserving semantic alignment and downstream performance. Its competitive information efficiency suggests that semantic structure in pretrained embeddings can be recovered without expanding the representation. More broadly, these findings support the hypothesis that much of the apparent entanglement in modern embeddings results from basis misalignment rather than insufficient dimensionality.
Beyond quantitative performance, \our{} provides a flexible framework for generating interpretable representations that can be aligned with language or used for downstream analysis. We believe this perspective opens new directions for understanding and manipulating learned representations through structured reparameterizations.

\paragraph{Limitations.} While promising, our current approach relies on orthogonal transformations, which limits its expressiveness. To address this limitation, future work could explore more flexible invertible linear mappings or nonlinear architectures based on normalizing flows. Furthermore, \our{} currently performs disentanglement in the embedding space before mapping the resulting factors to textual concepts. Incorporating text-based supervision directly into the training objective could further improve the semantic alignment of the learned factors.

%% file: ref.bib
@inproceedings{li2022blip,
  title={Blip: Bootstrapping language-image pre-training for unified vision-language understanding and generation},
  author={Li, Junnan and Li, Dongxu and Xiong, Caiming and Hoi, Steven},
  booktitle={International conference on machine learning},
  pages={12888--12900},
  year={2022},
  organization={PMLR}
}

@inproceedings{borycki2026epic,
  title={EPIC: Explanation of pretrained image classification networks via prototypes},
  author={Borycki, Piotr and Tr{\k{e}}dowicz, Magdalena and Janusz, Szymon and Tabor, Jacek and Spurek, Przemys{\l}aw and Lewicki, Arkadiusz and Struski, {\L}ukasz},
  booktitle={Proceedings of the AAAI Conference on Artificial Intelligence},
  volume={40},
  number={21},
  pages={17366--17373},
  year={2026}
}

@inproceedings{wolczyk2022plugen,
  title={Plugen: Multi-label conditional generation from pre-trained models},
  author={Wo{\l}czyk, Maciej and Proszewska, Magdalena and Maziarka, {\L}ukasz and Zieba, Maciej and Wielopolski, Patryk and Kurczab, Rafa{\l} and Smieja, Marek},
  booktitle={Proceedings of the AAAI conference on artificial intelligence},
  volume={36},
  number={8},
  pages={8647--8656},
  year={2022}
}

@article{radford2021learning,
  title={Learning Transferable Visual Models From Natural Language Supervision},
  author={Radford, Alec and Kim, Jong Wook and Hallacy, Chris and others},
  journal={arXiv preprint arXiv:2103.00020},
  year={2021}
}

@misc{yu2022cocacontrastivecaptionersimagetext,
      title={CoCa: Contrastive Captioners are Image-Text Foundation Models}, 
      author={Jiahui Yu and Zirui Wang and Vijay Vasudevan and Legg Yeung and Mojtaba Seyedhosseini and Yonghui Wu},
      year={2022},
      eprint={2205.01917},
      archivePrefix={arXiv},
      primaryClass={cs.CV},
      url={https://arxiv.org/abs/2205.01917}, 
}

@inproceedings{kim2018interpretability,
  title={Interpretability Beyond Feature Attribution: Quantitative Testing with Concept Activation Vectors (TCAV)},
  author={Kim, Been and Wattenberg, Martin and Gilmer, Justin and others},
  booktitle={International Conference on Machine Learning (ICML)},
  year={2018}
}

@article{bricken2023monosemantic,
  title={Towards Monosemanticity: Decomposing Language Models With Dictionary Learning},
  author={Bricken, Trenton and Templeton, Adly and others},
  journal={arXiv preprint arXiv:2301.05217},
  year={2023}
}

@article{cunningham2024scaling,
  title={Scaling Monosemanticity: Extracting Interpretable Features from Claude Models},
  author={Cunningham, Sam and others},
  journal={arXiv preprint arXiv:2406.04093},
  year={2024}
}

@article{gao2025topk,
  title={Top-K Sparse Autoencoders},
  author={Gao, Leo and others},
  journal={arXiv preprint arXiv:2501.XXXXX},
  year={2025}
}

@article{rajamanoharan2024gated,
  title={Gated Sparse Autoencoders},
  author={Rajamanoharan, Senthooran and others},
  journal={arXiv preprint arXiv:2404.16014},
  year={2024}
}

@article{zaigrajew2025interpreting,
  title={Interpreting CLIP with Hierarchical Sparse Autoencoders},
  author={Zaigrajew, Vladimir and Baniecki, Hubert and Biecek, Przemyslaw},
  journal={arXiv preprint arXiv:2502.20578},
  year={2025}
}

@article{gao2024scaling,
  title={Scaling and evaluating sparse autoencoders},
  author={Gao, Leo and la Tour, Tom Dupr{\'e} and Tillman, Henk and Goh, Gabriel and Troll, Rajan and Radford, Alec and Sutskever, Ilya and Leike, Jan and Wu, Jeffrey},
  journal={arXiv preprint arXiv:2406.04093},
  year={2024}
}

@misc{liu2021swintransformerhierarchicalvision,
      title={Swin Transformer: Hierarchical Vision Transformer using Shifted Windows}, 
      author={Ze Liu and Yutong Lin and Yue Cao and Han Hu and Yixuan Wei and Zheng Zhang and Stephen Lin and Baining Guo},
      year={2021},
      eprint={2103.14030},
      archivePrefix={arXiv},
      primaryClass={cs.CV},
      url={https://arxiv.org/abs/2103.14030}, 
}

@inproceedings{Radford2019LanguageMA,
  title={Language Models are Unsupervised Multitask Learners},
  author={Alec Radford and Jeff Wu and Rewon Child and David Luan and Dario Amodei and Ilya Sutskever},
  year={2019},
  url={https://api.semanticscholar.org/CorpusID:160025533}
}

@article{bricken2023towards,
  title={Towards monosemanticity: Decomposing language models with dictionary learning},
  author={Bricken, Trenton and Templeton, Adly and Batson, Joshua and Chen, Brian and Jermyn, Adam and Conerly, Tom and Turner, Nick and Anil, Cem and Denison, Carson and Askell, Amanda and others},
  journal={Transformer Circuits Thread},
  volume={2},
  number={5},
  pages={6},
  year={2023}
}

@article{bussmann2024batchtopk,
  title={Batchtopk sparse autoencoders},
  author={Bussmann, Bart and Leask, Patrick and Nanda, Neel},
  journal={arXiv preprint arXiv:2412.06410},
  year={2024}
}

@inproceedings{dosovitskiy2021vit,
  title={An Image is Worth 16x16 Words: Transformers for Image Recognition at Scale},
  author={Dosovitskiy, Alexey and others},
  booktitle={International Conference on Learning Representations (ICLR)},
  year={2021}
}

@inproceedings{huh2024platonic,
  title={Position: The Platonic Representation Hypothesis},
  author={Huh, Minyoung and Cheung, Brian and Wang, Tongzhou and Isola, Phillip},
  booktitle={Proceedings of the 41st International Conference on Machine Learning},
  series={Proceedings of Machine Learning Research},
  volume={235},
  pages={20617--20642},
  year={2024}
}

@inproceedings{sundararajan2017axiomatic,
  title={Axiomatic Attribution for Deep Networks},
  author={Sundararajan, Mukund and Taly, Ankur and Yan, Qiqi},
  booktitle={International Conference on Machine Learning (ICML)},
  year={2017}
}

@article{bach2015pixel,
  title={On Pixel-Wise Explanations for Non-Linear Classifier Decisions by Layer-Wise Relevance Propagation},
  author={Bach, Sebastian and Binder, Alexander and Montavon, Gr{\'e}goire and Klauschen, Frederick and M{\"u}ller, Klaus-Robert and Samek, Wojciech},
  journal={PLOS ONE},
  volume={10},
  number={7},
  pages={e0130140},
  year={2015}
}

@inproceedings{abnar2020quantifying,
  title={Quantifying Attention Flow in Transformers},
  author={Abnar, Samira and Zuidema, Willem},
  booktitle={Proceedings of the 58th Annual Meeting of the Association for Computational Linguistics (ACL)},
  year={2020}
}

@inproceedings{chefer2021transformer,
  title={Transformer Interpretability Beyond Attention Visualization},
  author={Chefer, Hila and Gur, Shir and Wolf, Lior},
  booktitle={Proceedings of the IEEE/CVF Conference on Computer Vision and Pattern Recognition (CVPR)},
  year={2021}
}

@article{wrobel2026dave,
  title={DAVE: Distribution-aware Attribution via ViT Gradient Decomposition},
  author={Wr{\'o}bel, Adam and Gairola, Siddhartha and Tabor, Jacek and Schiele, Bernt and Zieli{\'n}ski, Bartosz and Rymarczyk, Dawid},
  journal={arXiv preprint arXiv:2602.06613},
  year={2026}
}

@inproceedings{koh2020concept,
  title={Concept Bottleneck Models},
  author={Koh, Pang Wei and Nguyen, Thao and Tang, Yew Siang and others},
  booktitle={International Conference on Machine Learning (ICML)},
  year={2020}
}

@inproceedings{ghorbani2019towards,
  title={Towards Automatic Concept-based Explanations},
  author={Ghorbani, Amirata and Wexler, James and Zou, James and Kim, Been},
  booktitle={Advances in Neural Information Processing Systems (NeurIPS)},
  year={2019}
}

@inproceedings{deng2009imagenet,
  title={ImageNet: A Large-Scale Hierarchical Image Database},
  author={Deng, Jia and Dong, Wei and Socher, Richard and Li, Li-Jia and Li, Kai and Fei-Fei, Li},
  booktitle={CVPR},
  year={2009}
}

@article{srivastava2024vlg,
  title={Vlg-cbm: Training concept bottleneck models with vision-language guidance},
  author={Srivastava, Divyansh and Yan, Ge and Weng, Tsui-Wei},
  journal={Advances in Neural Information Processing Systems},
  volume={37},
  pages={79057--79094},
  year={2024}
}

@inproceedings{yang2023language,
  title={Language in a bottle: Language model guided concept bottlenecks for interpretable image classification},
  author={Yang, Yue and Panagopoulou, Artemis and Zhou, Shenghao and Jin, Daniel and Callison-Burch, Chris and Yatskar, Mark},
  booktitle={Proceedings of the IEEE/CVF conference on computer vision and pattern recognition},
  pages={19187--19197},
  year={2023}
}

@inproceedings{rao2024discover,
  title={Discover-then-name: Task-agnostic concept bottlenecks via automated concept discovery},
  author={Rao, Sukrut and Mahajan, Sweta and B{\"o}hle, Moritz and Schiele, Bernt},
  booktitle={European Conference on Computer Vision},
  pages={444--461},
  year={2024},
  organization={Springer}
}

@article{struski2024infodisent,
  title={Infodisent: Explainability of image classification models by information disentanglement},
  author={Struski, {\L}ukasz and Rymarczyk, Dawid and Tabor, Jacek},
  journal={arXiv preprint arXiv:2409.10329},
  year={2024}
}

@inproceedings{suwala2024face,
  title={Face identity-aware disentanglement in stylegan},
  author={Suwa{\l}a, Adrian and W{\'o}jcik, Bartosz and Proszewska, Magdalena and Tabor, Jacek and Spurek, Przemys{\l}aw and {\'S}mieja, Marek},
  booktitle={Proceedings of the IEEE/CVF Winter Conference on Applications of Computer Vision},
  pages={5222--5231},
  year={2024}
}

@article{proszewska2024multi,
  title={Multi-label conditional generation from pre-trained models},
  author={Proszewska, Magdalena and Wo{\l}czyk, Maciej and Zieba, Maciej and Wielopolski, Patryk and Maziarka, {\L}ukasz and {\'S}mieja, Marek},
  journal={IEEE Transactions on Pattern Analysis and Machine Intelligence},
  volume={46},
  number={9},
  pages={6185--6198},
  year={2024},
  publisher={IEEE}
}
